\documentclass[11pt]{article}

\usepackage[a4paper,margin=2.5cm]{geometry}
\usepackage[T1]{fontenc}
\usepackage{lmodern}
\usepackage{microtype}
\usepackage{amsmath,amssymb}
\usepackage{graphicx}
\usepackage{booktabs}
\usepackage{array}
\newcolumntype{L}[1]{>{\raggedright\arraybackslash}p{#1}}
\newcolumntype{C}[1]{>{\centering\arraybackslash}p{#1}}
\usepackage{xcolor}
\usepackage{fvextra}
\fvset{fontsize=\scriptsize, breaklines=true, breakafter=\,, frame=lines, framesep=1ex}
\usepackage{needspace}
\usepackage{tikz}
\usetikzlibrary{positioning,arrows.meta,fit,backgrounds,calc}
\usepackage[numbers,sort&compress]{natbib}
\usepackage[colorlinks=true,allcolors=blue!50!black]{hyperref}
\usepackage[capitalise,noabbrev]{cleveref}

\title{AI agents can redesign engineering parts with more than a thousand CAD parameters and
nonlinear simulations}
\author{Alexandra Starostina\textsuperscript{1}, Alexander Hermann\textsuperscript{2}, Mahish Guru\textsuperscript{1},\\ Kartik Bali\textsuperscript{1},
   Roland Aydin\textsuperscript{3,4} \& Christian C. Cyron\textsuperscript{1,2}\\[2ex]
  {\small
  \begin{tabular}{@{}c@{}}
    \textsuperscript{1}Helmholtz-Zentrum Hereon, Institute of Material Systems Modeling,\\
    Max-Planck-Straße 1, 21502 Geesthacht, Germany\\[0.8ex]
    \textsuperscript{2}Hamburg University of Technology,\\
    Institute for Artificial Intelligence and Simulation in Mechanics,\\
    Eißendorfer Str.\ 42, 21073 Hamburg, Germany\\[0.8ex]
    \textsuperscript{3}Saarland University,\\
    Research Group Data Driven Simulation and Analysis in Material Science,\\
    Campus, Building D3.3, 66123 Saarbrücken, Germany\\[0.8ex]
    \textsuperscript{4}German Research Center for Artificial Intelligence,\\
    Trippstadter Str.\ 122, 67663 Kaiserslautern, Germany
  \end{tabular}}}
\date{}

\begin{document}

\maketitle

\begin{abstract}
Simulation-driven design of real engineering components has to meet three conditions at once. It runs on few
solves, since a nonlinear finite-element simulation of a 3D part takes hours; its design space
has many values, the hundreds to thousands of values of the part's CAD model; and a better design often needs
a redesign, coordinated changes of the part's proportions rather than local polishing of its surface. Automated
optimisers meet at most two of these conditions. Classical optimisers, Bayesian or derivative-free, need a person
to cut the design down to a few parameters, and on all the values they need far more solves; industrial shape optimisers
move every surface node of the mesh in one solve per cycle, but only in small steps that polish rather than
redesign. We ask whether a general-purpose large language model (LLM) can meet all three by joining what it knows with
what the simulation shows. GPT-6 Astra edits a magnesium stent through a CAD geometry parametrisation of
about 1\,650 values, with a budget of 15 simulations; every proposal is built in CAD, meshed and simulated with
large-deformation plasticity and contact, and the LLM reads the full results of every solve. Within that
budget it lowered the objective, a high percentile of the strains in the stent's bends, by 32\%; on the same
budget, the industrial shape optimiser Tosca improved the original design by a tenth and then stopped
improving. An earlier GPT-6 Astra design, 25\% lower in the same objective, was checked in an
independent, higher-fidelity model of the full deployment with crimping and damage: it was 14\% less strained
than the original after the full chain, and its most damaged spot began to fail later and carried 63\% less
damage at full expansion. LLMs can thus join their learned knowledge with simulation data to
adjust more than a thousand design values and reach valid designs with real improvements.
\end{abstract}

\section{Introduction}\label{sec:intro}

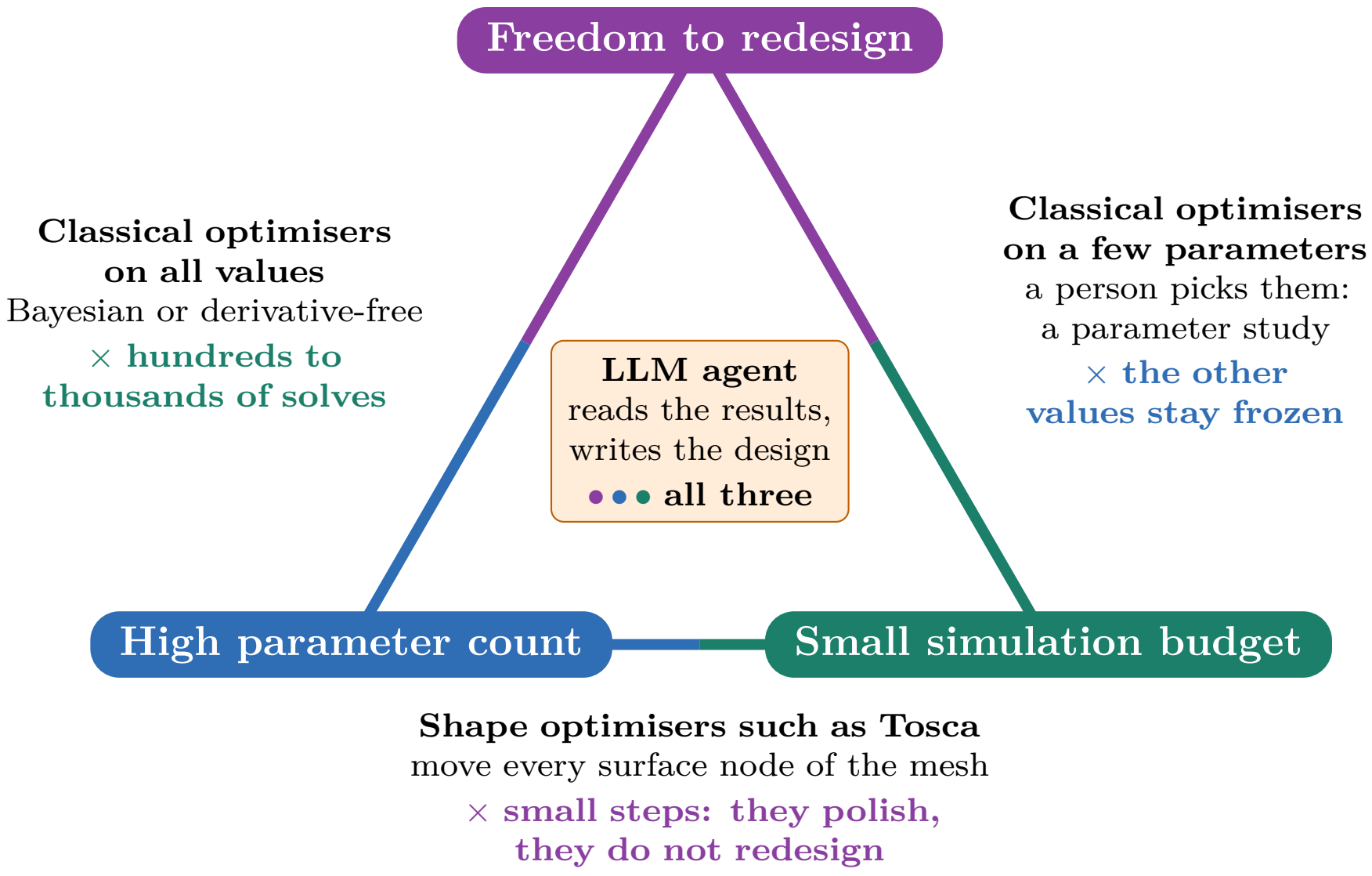
\begin{figure}[t]
  \centering
  \definecolor{condRedesign}{HTML}{8A3FA0}
\definecolor{condValues}{HTML}{2F6DB5}
\definecolor{condSolves}{HTML}{1B7F6A}
\begin{tikzpicture}[
    font=\footnotesize,
    pill/.style={rounded corners=7pt, text=white, font=\small\bfseries, inner xsep=7pt, inner ysep=3.5pt},
    family/.style={align=center, font=\scriptsize, inner sep=2pt, text width=44mm,
      execute at begin node={\hyphenpenalty=10000\exhyphenpenalty=10000}},
    side/.style={line width=2.6pt}
  ]
  \coordinate (R) at (90:3.4);
  \coordinate (M) at (210:3.4);
  \coordinate (F) at (330:3.4);
  \draw[side, condRedesign] (R) -- ($(R)!0.5!(M)$);
  \draw[side, condValues] ($(R)!0.5!(M)$) -- (M);
  \draw[side, condRedesign] (R) -- ($(R)!0.5!(F)$);
  \draw[side, condSolves] ($(R)!0.5!(F)$) -- (F);
  \draw[side, condValues] (M) -- ($(M)!0.5!(F)$);
  \draw[side, condSolves] ($(M)!0.5!(F)$) -- (F);
  \node[pill, fill=condRedesign] (r) at (R) {Freedom to redesign};
  \node[pill, fill=condValues] (m) at (M) {High parameter count};
  \node[pill, fill=condSolves] (f) at (F) {Small simulation budget};
  \node[family, anchor=east] at ($(R)!0.5!(M) + (-0.35, 0.25)$)
    {\textbf{Classical optimisers on all values}\\Bayesian or derivative-free\\[0.4ex]
     \textcolor{condSolves}{\textbf{$\times$ hundreds to thousands of solves}}};
  \node[family, anchor=west] at ($(R)!0.5!(F) + (0.35, 0.25)$)
    {\textbf{Classical optimisers on a few parameters}\\a person picks them: a parameter study\\[0.4ex]
     \textcolor{condValues}{\textbf{$\times$ the other values stay frozen}}};
  \node[family, anchor=north, text width=60mm] at ($(M)!0.5!(F) + (0, -0.5)$)
    {\textbf{Shape optimisers such as Tosca}\\move every surface node of the mesh\\[0.4ex]
     \textcolor{condRedesign}{\textbf{$\times$ small steps: they polish, they do not redesign}}};
  \node[align=center, draw=orange!75!black, fill=orange!15, rounded corners=3pt, inner sep=4pt, font=\scriptsize]
    at (0, 0.1) {\textbf{LLM agent}\\reads the results,\\writes the design\\[0.4ex]
    \textcolor{condRedesign}{$\bullet$}\,\textcolor{condValues}{$\bullet$}\,\textcolor{condSolves}{$\bullet$}\
    \textbf{all three}};
\end{tikzpicture}
  \caption{Three conditions of simulation-driven design, in colour at the corners. Each family of automated
  optimisers meets the two conditions its side joins and gives up the third, marked in that condition's colour;
  the agent of this paper meets all three.}
  \label{fig:conditions}
\end{figure}

Engineers improve a component by simulating it, reading the results and changing the design. For a
nonlinear three-dimensional part, three conditions shape that work (\cref{fig:conditions}). \emph{Small
simulation budget}: each finite-element simulation takes hours, so a design study runs on tens of them.
\emph{High parameter count}: the CAD model has hundreds to thousands of values that could change, from the
proportions of the part to the shape of every local feature. \emph{Freedom to redesign}: a better design often
needs coordinated changes of the proportions, not only local polishing of the surface. Automated optimisers meet at most two of the three.
Classical optimisers, derivative-free or Bayesian~\citep{powell1994cobyla,jones1998ego}, redesign within tens
of solves once a person has cut the design down to a few parameters, the usual parameter study. On all the
values they would need more solves than the whole budget just to start: a simplex method needs one solve per
value plus one, and high-dimensional Bayesian optimisers are designed and tested for budgets of hundreds to
thousands of evaluations~\citep{eriksson2019turbo,eriksson2021saasbo,hvarfner2024vanilla}. Industrial shape
optimisers such as Tosca move every surface node of the finite-element mesh in one solve per cycle, guided by
a stress controller or by sensitivities the solver computes~\citep{bakhtiary1996caoss,tosca}. They take small
steps on a mesh rather than a CAD model, so they polish a design but do not redesign it. None of them reasons
about what an engineer reads off a simulation: where a design is strained, why, and which change would help.

Large language models (LLMs) bring physics knowledge, read rich results such as tables, fields and plots,
write code to analyse them and act through tools. They have been placed in design loops with finite-element
or flow feedback~\citep{jadhav2024mechanicaldesigner,zhang2024llmpso,guo2025mdoagent}, and on engineering
simulators their advantage shows at small
budgets~\citep{wang2026calibrationagent,buonanno2026autoenergydesign}. \Citet{takehi2026weightspace}
showed that a frontier LLM can read the weights of a neural network and edit them much as a programmer
edits code, without gradients. We ask the same of a design: working through the CAD model of a real part with
a budget of fifteen simulations, can a general-purpose LLM reach a real improvement by using both the
simulation results and its own knowledge?

We give GPT-6 Astra a balloon-expandable magnesium stent. Expanding
it is a large-deformation elasto-plastic problem with contact, and the strain concentrates in the crowns, the
bends of its zigzag struts. A redesign must lower that strain within tight limits: the expanded stent must
stay as wide inside (its lumen), shorten no more and use no more metal. The agent writes design files that set
up to about 1\,650 values of a geometry parametrisation, from the stent's shared proportions down to smooth
bumps on every crown edge. Every proposal is built in CAD, meshed afresh and simulated, and the full fields of
every solve are available in a sandbox where the agent analyses them with its own code. The agent ran on the
stent twice: first in pilot sessions that helped to develop the loop, over which the parametrisation grew in
stages at its request, and then in a clean run that tested the finished loop, with the complete parametrisation
from the first call and nothing from the pilot sessions. In the
clean run, within fifteen simulations of about two hours each on one CPU, it lowered the objective, a high
percentile of the crown strains, by 32\% (\cref{fig:runs}b), for US\$19 of LLM calls; the gain is far
larger than the change in score when the same design is built and meshed again, and it holds on every other
measure we read. The agent uses the results, not only its prior knowledge: when the per-crown results it had
just received were shuffled between crowns, it expected the highest strains where the shuffled numbers put them,
and it no longer aimed its changes at the crowns that the simulation had actually found most strained
(\cref{sec:reading}). The finite-element model we optimise on is
a reduced one, the expansion alone. One of the improved designs was checked in an
independent, higher-fidelity model of the full deployment with crimping and damage, which the agent never saw
and which takes 17 to 39 hours per run; there, too, it was less strained than the original, and its most
damaged spot began to fail later and carried less damage (\cref{sec:fullcycle}).

Among automated methods, only an industrial shape optimiser can compete at this level of detail and budget,
so we ran Tosca from the same design with the same fifteen simulations. It polishes rather than redesigns:
from the original design it gained a tenth and then stopped improving, but started from a design the agent
had already redesigned, it polished that design well (\cref{sec:tosca}). Two smaller problems add evidence
(\cref{sec:ladder}). On Tosca's own bundled example, a plate with a hole described by up to 26 values, four
LLMs beat Tosca as shipped in 9 of 12 runs. On a single crown with four values, fewer than the
solves, a good Bayesian optimiser is a real competitor, and the LLMs' lead depends on the run.

\paragraph{Contributions.}
\begin{enumerate}
  \item A demonstration that a general-purpose LLM can redesign a nonlinear three-dimensional component
        through about 1\,650 CAD values within fifteen simulations, lowering its objective by a third for
        US\$19 of LLM calls (\cref{sec:headline}); to our knowledge, the first stent designed by an LLM
        agent. The gain achieved by the LLM design holds in an independent,
        higher-fidelity model of the full deployment (\cref{sec:fullcycle}).
  \item Evidence on how the agent approaches the problem, and why it works
        (\cref{sec:how}): it picks its own structure, giving whole families of crowns the same shape and
        correcting single crowns only at the end, and the scale of each move, concepts first and
        polishing last. Its first idea comes from prior knowledge; after that it uses the results: its
        predictions sharpen as it reads them, and when its per-crown results are shuffled between crowns, it
        no longer aims its changes at the crowns that the simulation actually found most strained.
  \item To our knowledge, the first comparison of an LLM design agent with an industrial shape optimiser:
        Tosca on the stent, from the original design and from the agent's first fix (\cref{sec:tosca}), and
        Tosca's own example run as shipped (\cref{app:plate}).
  \item A ladder of three problems at thirteen to fifteen simulations (\cref{sec:ladder}), in which the values
        set the part's geometry in increasing detail: four for a single crown (\cref{app:crown}); up to 26 for Tosca's plate with a hole (\cref{app:plate}); and about 1\,650 for the stent (\cref{sec:stent}). Where the values outnumber the solves, the agent leads; with fewer values than solves, a good classical
        optimiser catches up.
  \item Limits (\cref{sec:discussion}): the simulation set-up is still written by a person, because weaker
        LLMs accept wrong set-ups and report success. Moreover, the geometry of the CAD model has to be parametrised before the LLM can change it.
\end{enumerate}

\section{Related work}\label{sec:related}

\paragraph{LLMs as optimisers.}
LLMs can act as black-box optimisers: they propose new candidates from a prompt that lists
earlier candidates and their scores~\citep{yang2024opro}, serve as evolutionary operators, including on
low-dimensional engineering problems~\citep{brahmachary2024leo}, evolve programs against automated
evaluators~\citep{romeraparedes2024funsearch,novikov2025alphaevolve}, or strengthen and steer Bayesian
optimisation~\citep{liu2024llambo,brunzema2026sara}. A frontier LLM can also read the weights of neural
networks and edit them like a program, without gradients or training~\citep{takehi2026weightspace}; this work
started from that observation. In hyperparameter optimisation, classical optimisers beat pure LLM
search, while a hybrid that shares CMA-ES's internal state with an LLM did best~\citep{ferreira2026centaur}.
On engineering simulators, physics-labelled prompts gave an LLM a strong early lead in calibration, but the
lead was gone by about fifteen evaluations and varied widely between runs; a warm start handed to a
classical optimiser worked best~\citep{wang2026calibrationagent}. In these studies the LLM acts on a few
parameters or on code, guided by scores or short diagnostics. Our agent edits a CAD model of about
1\,650 values and reads the full field results of every solve.

\paragraph{LLM agents for simulation.}
LLM agents now write and run finite-element and CFD analyses from natural-language
specifications~\citep{ni2023mechagents,yue2026foamagent,hermann2026oasis}, including Abaqus
analyses~\citep{sarker2026abaqusagent} and complete geometry--mesh--simulation
pipelines~\citep{qi2025feagpt}. They automate set-up and execution; improving a design against a baseline
is not their aim.

\paragraph{LLMs in the design loop.}
\citet{jadhav2024mechanicaldesigner} let an LLM refine 2D truss designs from finite-element feedback, with
faster convergence and fewer evaluations than NSGA-II. LLM-PSO optimises parametric airfoil and
axisymmetric body shapes in laminar and Stokes flow~\citep{zhang2024llmpso}. An MDO agent turns
natural-language intent into parametric 3D CAD and optimises it through finite-element tool calls,
demonstrated on a gas-turbine blade, a machine-tool column and a heat sink~\citep{guo2025mdoagent}. Agents
also design electric motors with a surrogate and finite-element hybrid, beating finite-element-only search
on the same budget~\citep{han2026motordesign}; design building energy systems, coming within 1\% of the
best design in about 20 simulations of a full year where the best classical baseline needed about
190~\citep{buonanno2026autoenergydesign}; and shape cars with deep-learning
surrogates~\citep{elrefaie2025designagents}. In topology optimisation, LLMs set the continuation parameters
of SIMP~\citep{yang2026simpcontroller} or orchestrate a sensitivity-based solver~\citep{lin2026tomaster}.
Where these works state the size of the design that the LLM writes, it has 2 to 25
values~\citep{zhang2024llmpso,buonanno2026autoenergydesign}; in most of the others the LLM writes the
set-up, tunes a solver or steers a classical optimiser that does the search. Our agent's geometry
parametrisation has about 1\,650 values, some sixty times the largest of these, and from its third design on the agent set about
800 to 1\,200 of them (\cref{sec:how}).
To our knowledge, none of these compares an LLM agent with an industrial shape optimiser, works in the
large-deformation elasto-plastic regime with the CAD in the loop, or checks its result in an independent,
higher-fidelity model.

\paragraph{High-dimensional optimisation with few evaluations.}\label{sec:highdim}
Bayesian optimisation is the standard tool for expensive black-box functions. Its high-dimensional
variants fit local surrogate models~\citep{eriksson2019turbo}, assume that few of the variables
matter~\citep{eriksson2021saasbo}, or scale the prior with the dimension~\citep{hvarfner2024vanilla}. They
are designed for budgets of a few hundred to several thousand evaluations: even their smallest tests use 50
to 500, three to thirty-three times our fifteen, and their initial designs alone take 10 to 200 points.
SAASBO needed about 20 to 30 evaluations just to find the two variables that mattered among 100 in a
noise-free test. With fifteen evaluations in about 1\,650 dimensions, any surrogate model of the objective sees two
orders of magnitude fewer points than dimensions.
Grey-box Bayesian optimisation models intermediate outputs, such as a strain per crown, rather than the
score alone~\citep{astudillo2019composite}. It is the closest classical counterpart of reading the
results, but its surrogate models still grow with the number of design values. Gradient-based shape optimisation
handles many values with few solves when an adjoint sensitivity analysis supplies the gradient, the
derivatives with respect to all design values for about the cost of one more
simulation~\citep{jameson1988aerodynamic,tortorelli1994dsa}; but such sensitivities are not available off
the shelf for an objective like ours, through the whole chain from the CAD values to a new mesh, plasticity
and contact.

\paragraph{Stent design optimisation.}
\citet{wu2010stentshapeopt} optimised the shape of biodegradable magnesium stents with finite elements,
and a design from this line was tested in vitro and in vivo~\citep{chen2019invivo}. Other work combines
surrogates with multi-objective search on a few geometric
parameters~\citep{pant2011multiobjective,clune2014nurbs,wang2022krigingmgstent}, including Bayesian,
expected-improvement methods~\citep{ribeiro2021surrogate}. All of it relies on classical optimisers or
surrogates; we found no stent designed with an LLM.

\paragraph{Industrial non-parametric shape optimisation.}
Industrial optimisers such as Tosca build on optimality-criteria and controller-based
methods~\citep{bakhtiary1996caoss} and are established in industrial
practice~\citep{harzheim2006review,pedersen2006industrial}. In shape optimisation, Tosca moves the surface
nodes of the finite-element mesh directly, guided either by a stress target or by sensitivities of the
objective~\citep{simulia2019cae}, and returns a reshaped mesh rather than CAD.

\section{Overview of the LLM-driven optimisation loop}\label{sec:loop}

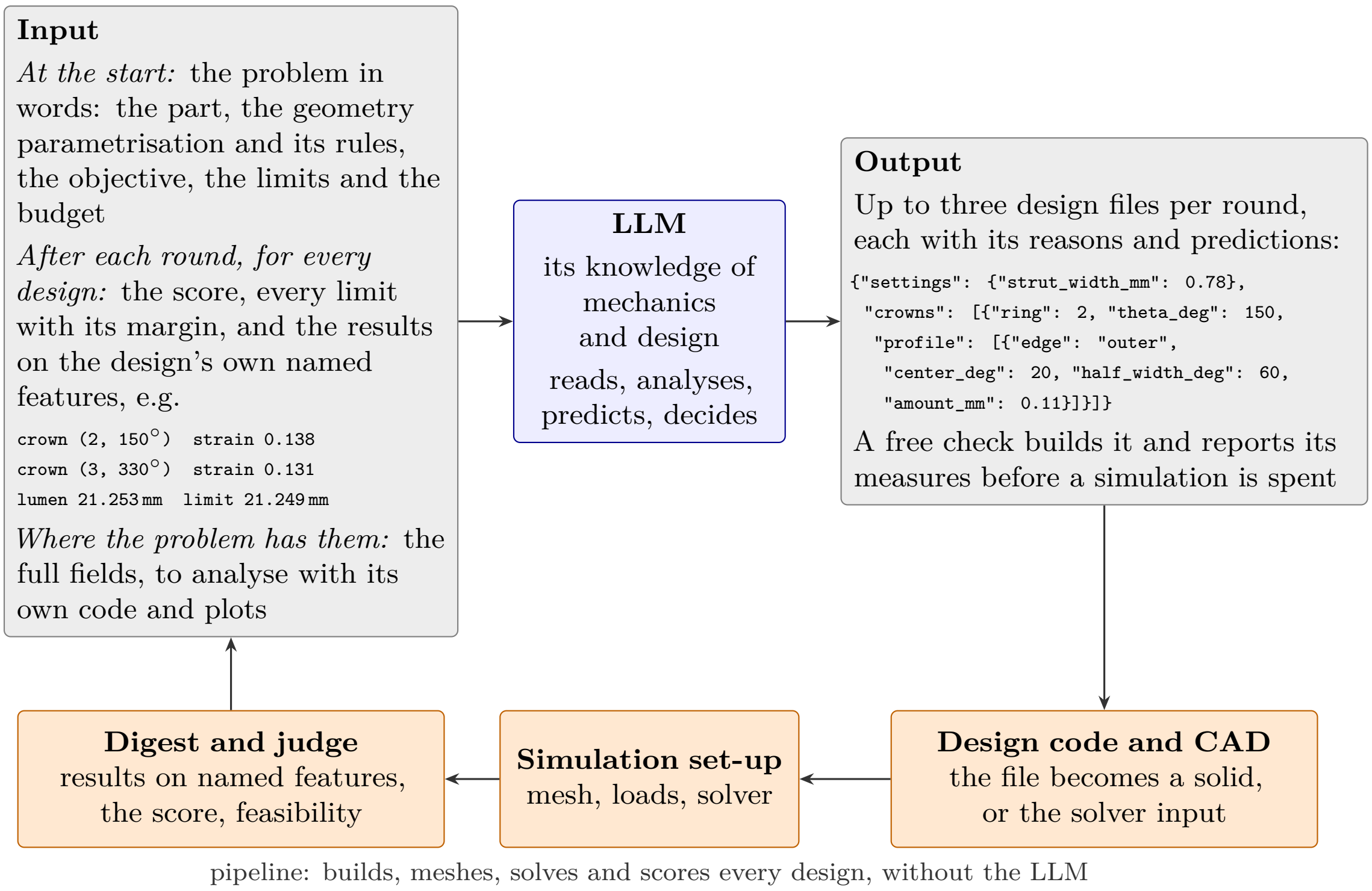
\begin{figure}[t]
  \centering
  \begin{tikzpicture}[
    font=\footnotesize,
    >={Stealth[length=4.5pt, width=3.6pt]},
    box/.style={rounded corners=2pt, inner sep=4pt, execute at begin node={\hyphenpenalty=10000\exhyphenpenalty=10000}},
    generic/.style={box, draw=black!50, fill=black!7, align=left},
    model/.style={box, draw=blue!55!black, fill=blue!7, align=center},
    supplied/.style={box, draw=orange!75!black, fill=orange!18, align=center, minimum height=15mm},
    flow/.style={->, semithick, black!80},
    lab/.style={font=\scriptsize, align=center, text=black!80, inner sep=1.5pt}
  ]
  \node[generic, text width=47mm] (receives) at (0, 0) {%
    \textbf{Input}\\[0.6ex]
    \textit{At the start:} the problem in words: the part, the geometry parametrisation and its rules, the
    objective, the limits and the budget\\[0.6ex]
    \textit{After each round, for every design:} the score, every limit with its margin, and the
    results on the design's own named features, e.g.\\[0.3ex]
    {\ttfamily\tiny crown (2, 150$^\circ$)\quad strain 0.138}\\[-0.4ex]
    {\ttfamily\tiny crown (3, 330$^\circ$)\quad strain 0.131}\\[-0.4ex]
    {\ttfamily\tiny lumen 21.253\,mm\quad limit 21.249\,mm}\\[0.6ex]
    \textit{Where the problem has them:} the full fields, to analyse with its own code and plots};
  \node[model, text width=27mm, right=6mm of receives] (llm) {%
    \textbf{LLM}\\[0.6ex]
    its knowledge of\\mechanics and design\\[0.6ex]
    reads, analyses,\\predicts, decides};
  \node[generic, text width=55mm, right=6mm of llm] (writes) {%
    \textbf{Output}\\[0.6ex]
    Up to three design files per round, each with its reasons and predictions:\\[0.3ex]
    {\ttfamily\tiny \{"settings": \{"strut\_width\_mm": 0.78\},}\\[-0.4ex]
    {\ttfamily\tiny \ "crowns": [\{"ring": 2, "theta\_deg": 150,}\\[-0.4ex]
    {\ttfamily\tiny \ \ "profile": [\{"edge": "outer",}\\[-0.4ex]
    {\ttfamily\tiny \ \ \ "center\_deg": 20, "half\_width\_deg": 60,}\\[-0.4ex]
    {\ttfamily\tiny \ \ \ "amount\_mm": 0.11\}]\}]\}}\\[0.6ex]
    A free check builds it and reports its measures before a simulation is spent};
  \draw[flow] (receives) -- (llm);
  \draw[flow] (llm) -- (writes);
  \node[supplied, text width=44mm, anchor=north] (digest) at ($(receives.south) + (0, -8mm)$)
    {\textbf{Digest and judge}\\results on named features,\\the score, feasibility};
  \node[supplied, text width=30mm, anchor=north] (simulate) at (llm.center |- digest.north)
    {\textbf{Simulation set-up}\\mesh, loads, solver};
  \node[supplied, text width=44mm, anchor=north] (geometry) at (writes.center |- digest.north)
    {\textbf{Design code and CAD}\\the file becomes a solid,\\or the solver input};
  \draw[flow] (writes) -- (geometry);
  \draw[flow] (geometry) -- (simulate);
  \draw[flow] (simulate) -- (digest);
  \draw[flow] (digest) -- (receives);
  \node[lab, anchor=north] at ($(simulate.south) + (0, -1mm)$)
    {pipeline: builds, meshes, solves and scores every design, without the LLM};
\end{tikzpicture}
  \caption{Overview of the LLM-driven optimisation loop, which runs in rounds of up to three designs. Top: the
  agent's input, the LLM with its knowledge, and the agent's output, the design files it writes. Bottom: the
  pipeline that turns each file into a scored simulation without the LLM; orange: the parts each problem
  supplies (\cref{tab:interface}). Example entries from the stent.}
  \label{fig:loop}
\end{figure}

Two questions decide whether an LLM can design through finite-element simulations at all: how the
results of a simulation reach it, and how it can specify a new geometry. \Cref{fig:loop} answers both. The
LLM reads results as text, organised by the design's own named features, and writes designs as short files
that set the values of a geometry parametrisation; building, meshing, solving and scoring run without it. The
loop runs in rounds, in each of which the LLM proposes up to three candidate designs at once
(\cref{sec:rounds}). From here on we call the LLM working in this loop, with its tools, the agent. We write LLM for the language
model itself, for instance when comparing LLMs; \emph{model} on its own always means a simulation model, such
as the finite-element model of a design. We describe the
loop on the stent of \cref{sec:stent}. The loop builds on FRAME, the Finite element Reasoning and AI-Agent Model
Engine~\citep{guru2026from}, and reuses parts of it.

\subsection{Simulation results as agent input}\label{sec:receives}
The agent works in a multi-turn session and keeps its messages, tool calls, results and reasoning across
turns. Its opening message (\cref{app:protocols}) states the problem in words: the part and its load case, the geometry
parametrisation with its ranges and rules, the objective, the limits and the budget, and the results of the
start design. After each round it receives, for every design it submitted, a digest: the score, every limit
with its margin, and tables of the results by the design's own named features. On the stent these are the
strains of every crown, link and ring, named by ring and angle as in the design files. Because the results and
the design files use the same names, a strain peak in the results points to the feature that can change it. The
agent also gets the full results: the fields of every node and element in every exported frame are available
in a Python sandbox with a small library to read them, and the agent writes its own analysis code there and
looks at the plots it draws. It never operates the solver or its graphical interface: everything reaches it as
text, tables and the images it draws itself. \Cref{tab:interface} lists the input and output for the stent
and for the two smaller problems of \cref{sec:ladder}.

\begin{table}[t]
  \centering
  \caption{Input and output of the agent for each problem. The plate runs through the same tools and
  geometry code as the stent; the crown ran in a lighter implementation of the same protocol, without sandbox
  or CAD (\cref{app:crown}).}
  \label{tab:interface}
  \small
  \begin{tabular}{@{}L{0.14\linewidth}L{0.28\linewidth}L{0.24\linewidth}L{0.24\linewidth}@{}}
    \toprule
    & Stent (\cref{sec:stent}) & Plate (\cref{app:plate}) & Crown (\cref{app:crown}) \\
    \midrule
    Agent writes & Shared settings, every crown and link, bumps on the crown edges; about 1\,650 values
    (\cref{tab:language}); outlines wrapped onto a tube & The hole's two semi-axes and up to eight bumps on
    its edge; up to 26 values; outline extruded & Three widths and a shift of the apex; four values, written
    into the solver input \\
    \addlinespace
    Free check & Rules, CAD build, measures & Rules, CAD build, measures & Rules on the widths \\
    \addlinespace
    Simulation & Balloon expansion and free recoil: large deformation, plasticity, contact; about two hours
    per solve & Tosca's example: linear elastic, two load cases; under a second & One crown opened and
    released: plasticity; one to two minutes \\
    \addlinespace
    It receives & Strain per crown, link and ring; lumen, shortening, volume & Stress along the edge of the
    hole, both load cases & Plastic strain along the crown, edge by edge \\
    \addlinespace
    Full results & Every node and element of every frame, in a Python sandbox & The same & The solver's
    output files \\
    \addlinespace
    Objective and limits & 90th percentile of the crown strains; lumen, shortening, volume, length, build
    rules & Peak von Mises stress at the hole; material area & Mean strain of the most strained 2\% of
    the metal; opening, shortening, volume \\
    \addlinespace
    Budget & 15 simulations, the start's included & 15, the start's included & 13, the start's included \\
    \addlinespace
    LLMs & GPT-6 Astra & GPT-6 Astra, Claude Fable~5.2, Kimi~K3.1~Pro, Mistral Large~4 & GPT-6 Astra,
    Mistral Large~4, Kimi~K3.1~Pro, DeepSeek~V4.1~Flash, Qwen~3.5~397B, Qwen~3.8 \\
    \bottomrule
  \end{tabular}
\end{table}

\subsection{Design files as agent output}\label{sec:language}
The agent writes design files: short JSON files that set the values of the problem's geometry
parametrisation. The parametrisation names the features of the part and what may change about them, with a
range for every value and rules that keep every design buildable. A file lists only what differs from shared
defaults, and anything left out takes the shared value. A parametrisation can be as small as a few shared
parameters, the form a classical optimiser needs, or reach down to single features of the part; the stent's
offers both, about 1\,650 values in all. The problem's design code, which the agent can read, turns a file
into a CAD solid (\cref{sec:stent-language}). Before it spends a simulation, the agent can check a file for
free: the check applies the rules, builds the geometry and reports its measures, such as the metal volume, and
what changed.

\subsection{Rounds of candidate designs}\label{sec:rounds}
A session runs in rounds. In each round the agent submits one to three candidate designs at once, each with its
reasons and its predictions of the main results. The pipeline simulates them, side by side as far as the
solver licences allow, and the agent receives the results of all of them together before it proposes the next
round. A candidate must pass the free
check and must not repeat a design already simulated; a round with a rejected candidate is returned as a whole
and costs nothing. Every candidate uses one simulation of the budget, which also counts the simulation of the
start design. The agent decides how many candidates to submit, up to three. Several candidates at once let it
compare alternatives on the same footing: different concepts in an early round, variants of one idea later, or
a single candidate for a last correction. They can also shorten a study when they run side by side; in the
clean run they mostly ran one after another, as the licences we share with colleagues rarely left room for two
(\cref{app:stent-model}). Its predictions show whether
it expects what the simulations then give (\cref{sec:reading}). The session ends when the budget is spent or
the agent finishes, returning its best design, a summary and the changes it wanted but could not express.

\subsection{Pipeline and protocol}\label{sec:pipeline}
A pipeline turns each submitted design into a scored simulation: it builds the geometry, meshes it, writes the
solver deck with the problem's set-up, solves it, exports the fields and computes the digest; on the stent it
meshes and solves with Abaqus~\citep{abaqus}. A scoring script, which we call the judge, then rates the design
against the objective and the limits. A problem enters the loop through four parts: the geometry
parametrisation with its code, the simulation set-up, the digest, and the objective with its limits. They are
the orange boxes of \cref{fig:loop}, where the digest shares a box with the judge that applies the objective
and the limits. The agent's tools and the rules below are the same for the stent and the plate; the crown ran
in a lighter implementation with a smaller budget (\cref{tab:interface}). On the stent we use GPT-6 Astra
through OpenRouter, at its highest reasoning effort in every session with tools (\cref{tab:protocols}). It ran
there twice: in pilot sessions that helped to develop
the loop, over which the geometry parametrisation grew at the agent's request, and in a clean run that tested
the finished loop, with the complete parametrisation from the first call and nothing from the pilot sessions
(\cref{sec:headline}).

Before the first LLM call of each run we fix its protocol (\cref{app:protocols}): the LLM, the opening message, the
tools, the budget, the cost caps and the questions the run is to answer. Every message, tool call, result and
cost is recorded; a run can be paused and resumed without repeating an LLM call or a simulation; and no human
intervenes during a session. On budgets of a few simulations, small choices can decide a comparison, so the
protocol and the judge close four loopholes. (i)~The objective, the limits and their tolerances are fixed
before the first call; a reading chosen after results exist, such as a looser tolerance, is reported beside
the fixed one and marked as such. (ii)~A design whose build, mesh or solve fails uses up its simulation and is
never feasible, so no method gains by proposing designs that break the pipeline. (iii)~Every limit is an
absolute value or the start design's own value, never a percentage, which a design can meet by changing the
denominator: a longer stent, for instance, shortens less in percent without any real improvement
(\cref{sec:reading}). (iv)~Two designs count as different only if their scores differ by more than the noise
of a rebuild, the change in score when the same design is built and meshed again (\cref{sec:noise}); likewise,
what would count as a result across repeated runs, such as a lead in at least two of three, is fixed with the
protocol.

\section{Stent design problem}\label{sec:stent}

This section explains in detail the headline nonlinear problem the agent worked on. We first describe what a stent does, why simulating its
deployment is demanding, and which reduced model we optimise on (\cref{sec:stent-sim}), and outline that
finite-element model (\cref{sec:stent-model}), with its details in \cref{app:stent-model}. We then state what
the redesign should achieve, as an objective and limits (\cref{sec:objective}). Last comes the geometry
parametrisation, through which the agent changes the stent's CAD model (\cref{sec:stent-language}).

\subsection{Stents and their simulation}\label{sec:stent-sim}
\begin{figure}[t]
  \centering
  \includegraphics[width=\linewidth]{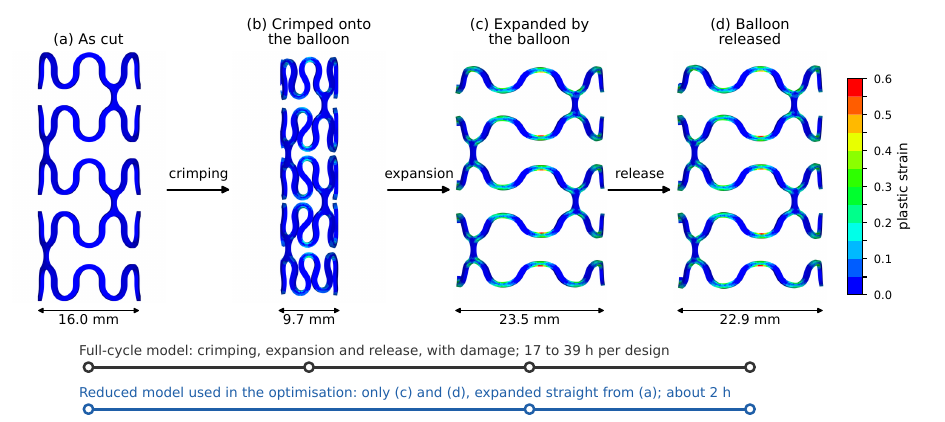}
  \caption{Deployment of the original design in the full-cycle model of \citet{steglich2026loadreversals},
  with the parameters of the magnesium alloy ZX10: (a)~as cut, (b)~crimped onto the balloon, (c)~expanded by
  the balloon and (d)~after the balloon's release. Side views at one scale, axis upright, front half only,
  each with the stent's outer diameter; the crimper and the balloon are not drawn. Colour: cumulative plastic
  strain, each element's largest integration-point value. Bottom: the steps each model runs, with its run time
  on one or two CPUs. The reduced model used in the optimisation (\cref{sec:stent-model}) covers only (c)
  and~(d): it leaves out crimping, and a rigid cylinder expands the as-cut stent until the cylinder is
  21.6\,mm across and is withdrawn.}
  \label{fig:deployment}
\end{figure}

A stent is a small tubular scaffold that holds a narrowed vessel open. A balloon-expandable stent is cut from a
thin-walled tube as rings of zigzag struts joined by links. It is crimped onto the balloon of a catheter,
carried to the narrowing and expanded by inflating the balloon, which bends the struts open at their crowns
until the stent presses against the vessel wall. When the balloon is deflated, the stent springs back a little
(its recoil) and must still hold the vessel open (\cref{fig:deployment}). The stent of this paper is cut from a
tube 16\,mm across, crimped to about 10\,mm and expanded to about 23\,mm, all as outer diameters. It is made of
magnesium, which
dissolves in the body over time, but magnesium alloys are much less ductile than the alloys of permanent
stents, so the strain that deployment puts into the crowns largely decides whether the stent is
damaged~\citep{wu2010stentshapeopt,steglich2026loadreversals}.

Simulating a deployment is demanding. The struts deform far beyond the elastic range, and the plastic strain
concentrates in the bends of the crowns; the stent touches the crimper, the balloon and itself; it springs back
after each step; and a magnesium alloy behaves differently in tension and compression and accumulates damage,
which matters because crimping and expansion bend the crowns in opposite directions. A full simulation
therefore follows the whole deployment, crimping, release, expansion and release, with a material model
calibrated on cyclic tests that captures load reversals and damage. \Citet{steglich2026loadreversals} built
such a model for this stent; a run took 17 hours for the original design and 39 hours for one of the agent's
designs, on one or two CPUs (\cref{app:stent-model}).

For the optimisation we use a reduced model that solves in about two hours on one CPU. It leaves out crimping:
it expands the stent from its as-cut state and lets it recoil, with plasticity but without load reversals or
damage (\cref{fig:deployment}, bottom). It keeps what makes the problem hard: large deformation, plasticity
concentrated in the crowns, contact and recoil, all strongly nonlinear in the design. Whether the reduced model
is a reliable surrogate for the full deployment is beyond the scope of this paper; we check one of the agent's
designs in the full model (\cref{sec:fullcycle}).

\subsection{Finite element model}\label{sec:stent-model}
The original design is the stent of \citet{steglich2026loadreversals}, derived from the Open Stent
Design~\citep{bonsignore_openstent}: five rings of twelve crowns each, joined by eight links that sit on
sixteen of the 60 crowns, cut from a tube 39\,mm long with a wall 0.6\,mm thick (\cref{fig:design}). In the
reduced model a rigid cylinder stands in for the balloon: it opens the stent from its as-cut state until the
cylinder is 21.6\,mm across, which takes the stent to an outer diameter of about 22.9\,mm, and is withdrawn,
so that the stent recoils freely. The magnesium alloy, Mg-4Y-3Gd, follows a plasticity model fitted to one
tensile test~\citep{steglich2025smartrim}, without damage and with the same response in tension and
compression; the full-cycle runs of \cref{fig:deployment,sec:fullcycle} use the parameters of another
magnesium alloy, ZX10. The two differ for a practical reason: ZX10's model is a custom material model,
calibrated on cyclic tests, so the optimisation uses a standard
plasticity model of an alloy whose test data are public. Three points on the middle ring, the supports, keep the
stent from sliding or turning as a whole while leaving it free to expand; they load the
six crowns next to them artificially (\cref{sec:objective}). Each design is meshed afresh with about 37\,000
brick elements and solved in Abaqus/Standard on one CPU core of Strand, the HPC cluster at Hereon, in
about two hours (1.9 to 2.3 hours in the clean run; \cref{app:stent-model}). Not modelled are crimping, load reversals, damage, the vessel,
balloon folding and manufacturing tolerances. The digest the agent receives lists, per crown, link and ring,
the strains and motions, together with the lumen, the shortening and the volume, which \cref{sec:objective}
defines. \Cref{app:stent-model} gives the material data, the supports, the contact and the mesh.

\subsection{Objective and limits}\label{sec:objective}
The redesign should lower the strain in the crowns, where the stent deforms most and where damage begins,
without giving up what the stent must do: hold the vessel open as wide as the original design, shorten no more
as it expands, and use no more metal. In agreement with established
practice~\citep{zhu2025ascending,jamshidian2026aaa,le2010stress,rockafellar2000cvar}, we set the objective as a
high percentile of the strain rather than its peak: $Q$ is the 90th percentile of the strains of the 60 crowns
after recoil; lower is better. A crown's strain is the mean of the five most strained elements that belong to
it, each element read at its most strained integration point (\cref{app:stent-model}). A single peak depends
strongly on the mesh and on the supports (\cref{app:stent-model}), whereas $Q$ is exceeded by only six crowns and
can be lowered only by bringing all the others below the new level; it follows the most strained crowns as a
group. Since $Q$ does not see how far those six lie above it, we also read the most strained crown and the most
strained element of each best design (\cref{tab:measures}). And since the supports load the six crowns next to
them artificially, $Q_{\mathrm{ex}}$, the objective without those six crowns, checks for that artefact. For the
original design $Q=0.1604$.

A design is feasible if (i) its lumen, the mean inner diameter after recoil, is at least the original's minus
0.01\,mm (21.249\,mm); (ii) its shortening, the as-cut length minus the length after recoil, is at most the
original's plus 0.03\,mm (1.3605\,mm); (iii) its metal volume, computed exactly from the CAD solid, is at most
the original's (229.19\,mm$^3$); (iv) its as-cut length is at most 41\,mm; and (v) it obeys the rules of the
geometry parametrisation (\cref{tab:language}). A design that cannot be built, meshed or solved counts as
infeasible, with $Q=0.25$. The limits are engineering choices rather than clinical criteria; shortening, or
foreshortening, for instance, is reported in regulatory testing without a pass--fail value~\citep{fda2010stents}.

The shortening limit has three readings, and every design is judged with each of them. The \emph{strict}
reading is the original's own value, 1.3305\,mm. The \emph{main} reading is (ii), with its tolerance of
0.03\,mm. That tolerance, like the 0.01\,mm on the lumen in (i), covers the noise of the set-up: moving only
the supports changed the original's lumen by 0.0009\,mm and its shortening by 0.028\,mm. A rebuild, with a new
CAD solid and a new mesh, scatters the shortening by more than its tolerance (\cref{app:noise}). The comparison
between methods therefore uses a third, the \emph{head-to-head} reading: the judge counts a design as breaking
the shortening limit only if it exceeds the main limit by more than the noise of a rebuild, 0.17\,mm, that is,
beyond 1.5305\,mm. The judge scores Tosca's designs against the same limits as far as they apply. Tosca's
designs are reshaped meshes, not files of the geometry parametrisation, so their metal volume is read from the
mesh, scaled to the start design's exact volume, and of the rules in (v) only the two that can be measured on a
mesh are checked, the width of the narrowest crown tip and the flank radii of the links. Tosca itself sees none
of these limits except the volume, which it holds at its start's value (\cref{sec:tosca-setup}).

\subsection{Geometry parametrisation}\label{sec:stent-language}
\begin{figure}[tbp]
  \centering
  \includegraphics[width=\linewidth]{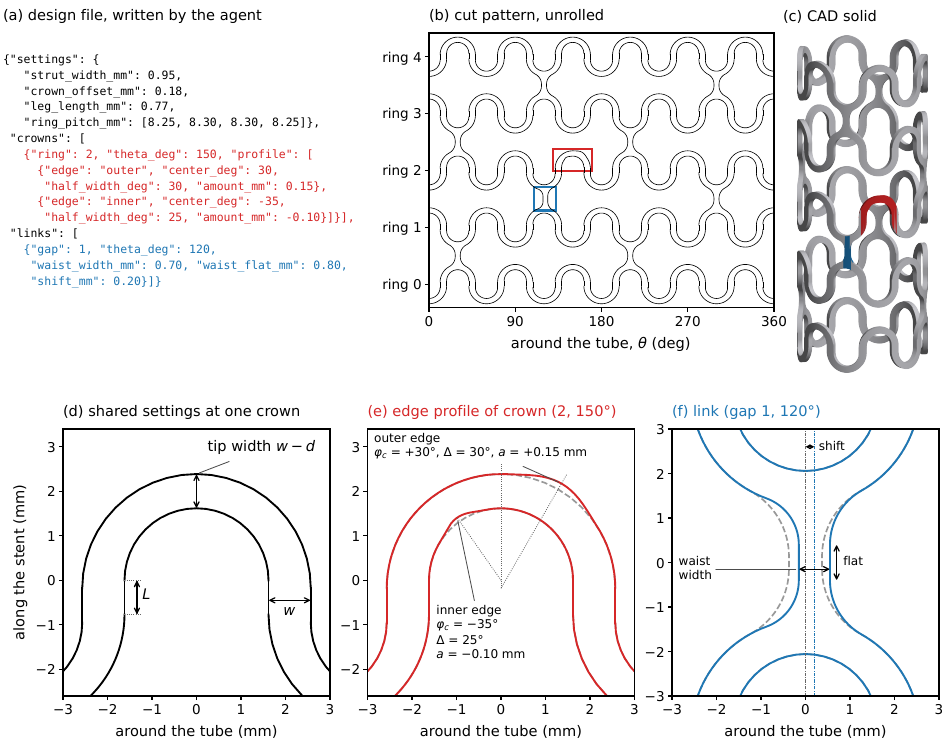}
  \caption{The stent's geometry parametrisation, from a file to the solid. (a)~An example design file as the agent
  writes it (its schema tag omitted): shared settings, where \texttt{crown\_offset\_mm} is the tip narrowing
  $d$, an entry for one crown with two bumps (red) and an entry for one link (blue). (b)~The cut pattern it describes, unrolled onto the plane at equal scale. Crowns
  are named (ring, $\theta$) and links (gap, $\theta$), where gap~$g$ lies between rings $g$ and $g+1$.
  (c)~The solid that FreeCAD builds from the file, with the same crown and link coloured. (d)~The shared
  settings at one crown: strut width $w$, tip width $w-d$ and leg length $L$. (e)~The two bumps of crown (2,
  $150^\circ$): material added on the outer edge and removed from the inner edge. (f)~The link in gap~1 at
  $120^\circ$, with a narrower and longer waist, shifted by 0.20\,mm around the tube. Dashed in (e) and (f):
  the same file without its crown and link entries.}
  \label{fig:design}
\end{figure}

The geometry parametrisation is how the agent changes the stent's CAD model (\cref{sec:language}). The agent writes values for the stent's named features, so the parametrisation decides
what it can change. It has three levels (\cref{tab:language}). The shared settings are the strut width $w$;
the tip narrowing $d$, which leaves a tip width $w-d$; the length $L$ of the straight legs; the ring pitch,
set once or gap by gap; and the waist width and flat of the links (\cref{fig:design}d). Each crown can take
its own width and tip narrowing and move along the axis, which lengthens its two legs when it moves outwards.
Each link can take its own waist width and flat, the length of its straight middle part, and move around the
tube (\cref{fig:design}f). Still at the level of the crown, edge profiles add up to six bumps to its inner or
outer edge, with the angle $\varphi$ measured from the crown's apex (\cref{fig:design}e). Crowns are named by
ring and angle around the tube, links by the gap they bridge and their angle (\cref{fig:design}b), the names
under which the agent also receives its results. The tube, the five rings of twelve crowns and the sixteen
crowns that carry the eight links are fixed. In all, a design can set up to about 1\,650 values, most of them
in edge profiles. A parameter study would use four of the shared settings: the strut width, the tip narrowing,
the leg length and the ring pitch.

The design code draws each crown as a U-bend between an
inner and an outer circular arc. The legs are common
tangents of neighbouring arcs, and each link is an hourglass whose flanks are arcs tangent to its crown and
to its waist, so every outline is tangent-continuous. At the angle $\varphi$ along an arc, a bump with centre
$\varphi_c$, half width $\Delta$ and amount $a$ moves the edge along its normal by $a\,(1-t^2)^3$, where
$t=(\varphi-\varphi_c)/\Delta$ and $|t|<1$, so the edge returns to its arc with continuous curvature; a
positive amount adds material and a negative amount removes it. The outlines lie on the unrolled outer
surface, with the axial position and the arc length as coordinates (\cref{fig:design}b). FreeCAD wraps them
onto the outer surface of the tube as B-spline curves, extends them radially to the inner surface as ruled
walls, splits the tube with all the walls and keeps the piece that holds the struts (\cref{fig:design}c).
The metal volume follows exactly from the outlines, and a build counts only if it gives a single valid solid
whose finely tessellated volume matches the exact volume to 0.05\% and which survives export to STEP and
re-import. Tests confirm that a bump changes only its own edge and adds the volume of its integral to within
2\%. The plate with a hole of \cref{sec:ladder} uses the same code. The parametrisation grew over pilot
sessions at the agent's request (\cref{sec:headline,sec:how}); in the clean run it is complete from the first call.

\begin{table}[t]
  \centering
  \caption{The stent's geometry parametrisation: the values a design file can set, and the rules every design
  must pass
  before it is simulated. Angles $\theta$ are measured around the tube, angles $\varphi$ from a crown's
  apex.}
  \label{tab:language}
  \small
  \begin{tabular}{@{}L{0.2\linewidth}L{0.47\linewidth}L{0.25\linewidth}@{}}
    \toprule
    Level & Control & Range \\
    \midrule
    Shared & Strut width $w$ & 0.76--1.10\,mm \\
    & Tip narrowing $d$ & 0--0.40\,mm \\
    & Leg length $L$ & 0.40--1.40\,mm \\
    & Ring pitch, once or per gap & 7.60--8.75\,mm \\
    & Link waist width & 0.57--1.20\,mm \\
    & Link flat & 0--2.0\,mm \\
    \addlinespace
    Per crown (60), & Width & 0.76--1.20\,mm \\
    named (ring, $\theta$) & Tip narrowing & $-0.30$ to 0.63\,mm \\
    & Extension along the axis & $-0.30$ to 0.40\,mm \\
    & Edge profile: up to six bumps, each on the inner or outer edge, with centre $\varphi_c$, half
    width $\Delta$ and amount $a$ & $|\varphi_c|\le80^\circ$, $5^\circ\le\Delta\le60^\circ$,
    $|a|\le0.30$\,mm \\
    \addlinespace
    Per link (8), & Waist width & 0.57--1.20\,mm \\
    named (gap, $\theta$) & Flat & 0--3.0\,mm \\
    & Shift around the tube & $-0.50$ to 0.50\,mm \\
    \addlinespace
    Fixed & \multicolumn{2}{L{0.74\linewidth}@{}}{The tube (15.99\,mm outer diameter, 0.593\,mm wall),
    five rings of twelve crowns, and the sixteen crowns that carry the eight links} \\
    \addlinespace
    Rules & \multicolumn{2}{L{0.74\linewidth}@{}}{Tip and local crown width at least 0.57\,mm; straight
    part of every leg at least 0.2\,mm; link flank radii at least 0.3\,mm; facing unlinked tips at least
    0.5\,mm apart; link attachments at least $3^\circ$ inside the crown arcs; bumps within $\pm85^\circ$ of
    the apex, ending $2^\circ$ before the legs and $5^\circ$ clear of a link attachment; as-cut length at
    most 41\,mm; metal volume at most the original's} \\
    \bottomrule
  \end{tabular}
\end{table}

\section{Results}\label{sec:results}

\subsection{Stent optimisation by the agent}\label{sec:headline}
\begin{figure}[t]
  \centering
  \includegraphics[width=0.8\linewidth]{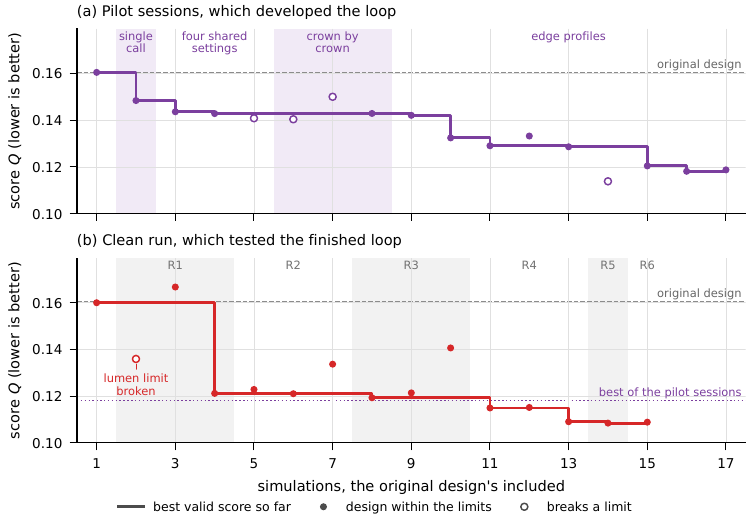}
  \caption{The agent's two runs on the stent: the score $Q$ of every design, the 90th percentile of its 60 crown
  strains (\cref{sec:objective}). The line is the best score among the valid designs so far. (a)~The pilot sessions, which developed the loop; shaded and named, the stages over
  which the geometry parametrisation grew at the agent's request. (b)~The clean run, with the complete
  parametrisation from the first call; shaded and labelled R1--R6, its six rounds (\cref{sec:rounds}), with the
  designs of each round in the order submitted. Dotted: the best
  score of the pilot sessions, which the clean run passes with its tenth design.}
  \label{fig:runs}
\end{figure}

The agent worked on the stent in two runs with different roles (\cref{fig:runs}). The pilot sessions were part of
developing the loop. The agent worked on the stent while the loop was still being built, and between its
sessions we extended the geometry parametrisation to cover the changes it asked for but could not yet express:
after a single call and a round on the four shared settings came crown-by-crown edits with the sandbox, and then
edge profiles (\cref{fig:runs}a). Along the way its designs lowered the objective $Q$, the 90th percentile of the crown strains,
from the original design's 0.1604 to 0.1181 with its fifteenth design, 26\% less.

The clean run then tested the finished loop. The agent received it complete from the first call, with the full
geometry parametrisation, the sandbox and the tools, and nothing from the pilot sessions. It lowered $Q$ from
0.1604 to 0.1085 with its thirteenth design, 32\% less, valid under every reading of the limits. It spent its
fourteen simulations in six rounds, of three,
three, three, three, one and one designs. Three quarters of the gain came from its first round, three designs
proposed together from the original design's results alone, the best of which cut $Q$ by 24\%
(\cref{fig:runs}b). The later rounds added
eight percentage points in steps of up to 0.006, and the last two were pure polishing. Thirteen of its
fourteen designs were valid under the head-to-head reading. The agent picked
its best design itself. Its LLM calls cost US\$19.44, against about two hours of cluster time for each of
its fifteen simulations. The cost was low because the run needed only 30 calls and the LLM wrote little,
building its design files with short Python programs: all it wrote cost US\$1.81, and the rest paid for
re-reading the conversation, which grew with every round. Only GPT-6 Astra ran on the stent; the two smaller
problems carry the comparison between LLMs (\cref{sec:ladder}).

\paragraph{Verification.}\label{sec:noise}
Every design gets its own CAD model and a new mesh, so even two practically identical designs do not score
exactly the same. To measure how much they differ, we rebuilt four designs with their struts a thousandth of a
millimetre thinner, a change too small to matter, and simulated them again: their scores changed by at most
0.003. We take 0.004, 2.5\% of the original design's score, as the noise of a rebuild: two designs count as
different only if their scores differ by more, which rebuilds of the same design rarely do
(\cref{app:noise}). The clean run's best design, rebuilt once more, scored 0.1088 instead of 0.1085 and still
met every limit. The agent's gains are an order of magnitude larger than this noise, 32\% in the clean run
and 26\% in the pilot sessions, and even the clean run's lead over the pilot sessions is more than twice it.

\subsection{Validation in the high-fidelity full-cycle model}\label{sec:fullcycle}
\begin{figure}[t]
  \centering
  \includegraphics[width=\linewidth]{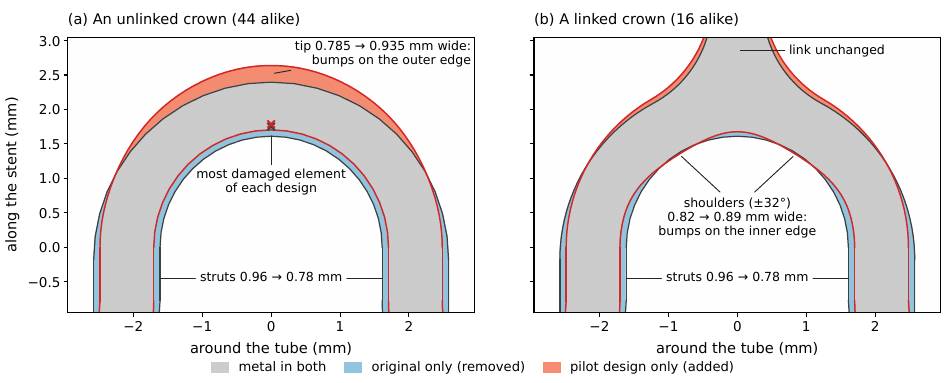}
  \caption{Geometric differences between the agent's design and the original, as cut. This design gives all
  crowns of a kind the same shape.
  Grey: metal in both designs; blue: only in the original; red: only in the pilot design. (a)~The 44
  unlinked crowns: thinner struts, a larger inner radius and, from two bumps on the outer edge, a wider tip;
  crosses: where each design's most damaged element sits in the full-cycle model (\cref{fig:damagefields}).
  (b)~The 16 linked crowns: thinner struts, with two bumps on the inner edge that widen the shoulders; the
  link is unchanged.}
  \label{fig:pilotgeometry}
\end{figure}

\begin{figure}[t]
  \centering
  \includegraphics[width=0.56\linewidth]{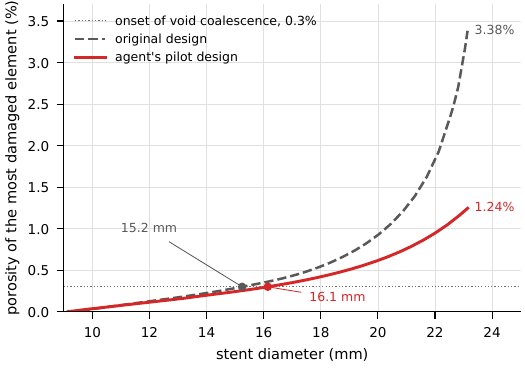}
  \caption{Damage of the agent's pilot design and the original in the full-cycle model of
  \citet{steglich2026loadreversals} with ZX10: both crimped, released and expanded, the two expansions from
  the same model set-up. The porosity of each design's most damaged element at full expansion, followed
  through the expansion. Dots: where it reaches 0.3\%, the onset of void coalescence; right: the values at
  full expansion.}
  \label{fig:fullcycle}
\end{figure}

The full-cycle model of \citet{steglich2026loadreversals} follows the whole deployment of this stent
(\cref{sec:stent-sim,fig:deployment}), with a Cazacu--Plunkett--Barlat yield criterion for the
tension--compression asymmetry of magnesium, kinematic hardening for load reversals and
Gurson--Tvergaard--Needleman damage, calibrated on cyclic tests.
One of the agent's pilot designs (\cref{fig:pilotgeometry}) was run through crimping and expansion with the parameters of ZX10, the alloy that in their study accumulates
damage fastest and fails before full deployment; both expansions come from the same model set-up
(\cref{fig:fullcycle}).

\begin{figure}[t]
  \centering
  \includegraphics[width=\linewidth]{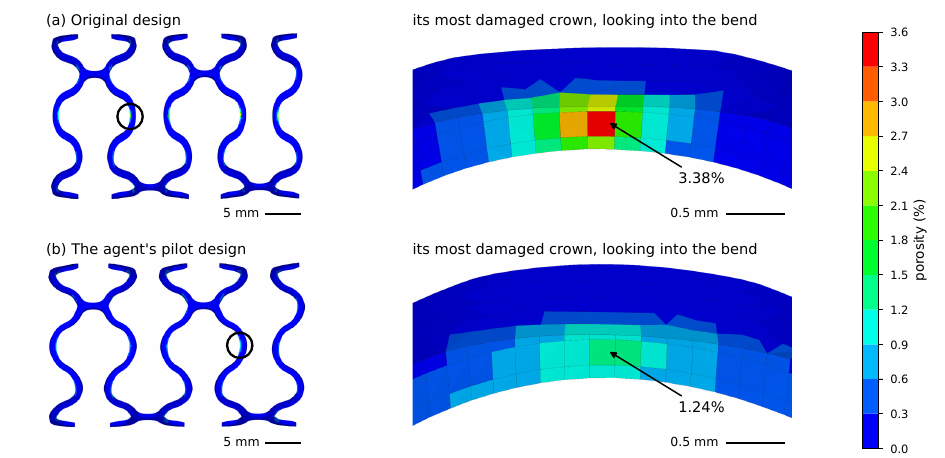}
  \caption{Damage fields at full expansion, in the model of \cref{fig:fullcycle}: (a)~the original,
  (b)~the agent's pilot design, each element coloured by its porosity, read as in \cref{fig:fullcycle}, on one scale; the darkest
  blue is below 0.3\%, the onset of void coalescence. Left: the half of the stent around its most damaged
  element, seen from outside; circle: that element's crown. Right: that crown, looking into its bend onto the
  cut wall where the element sits; arrow: the most damaged element and its porosity.}
  \label{fig:damagefields}
\end{figure}

After the full chain, the crowns of the agent's design were less strained throughout: the 90th
percentile of its crown strains was 14\% lower and
its most strained crown 24\% lower. Its most damaged spot reached the porosity at which void coalescence begins later, at a stent
diameter of 16.1\,mm instead of 15.2\,mm, and carried less damage at equal diameter: a third less at
20\,mm, about half at 22\,mm, and 63\% less at full expansion, 1.2\% against 3.4\%. In both designs that spot
sits on the inner edge of a crown's bend, on the cut wall (\cref{fig:damagefields}). There the two designs
differ most (\cref{fig:pilotgeometry}): the agent's design has thinner struts, 0.78 against 0.96\,mm, but
bumps on the outer edge widen the tips of its unlinked crowns from 0.785 to 0.935\,mm.

Our score targets the most
strained crowns, not the amount of strained metal: as the design spread the strain from its peaks, more of its
metal passed the coalescence porosity by full expansion, 8.6\% against 7.0\% of the stent. After crimping and release alone, the design was
about 12\% less strained than the original but needed about 16\% more crimping force, and facing crown tips
touched at full crimp.

\subsection{The agent's approach to the redesign}\label{sec:how}
This subsection asks three questions of the clean run: how much of the design the agent changed, whether
its changes redesign the stent or merely reshape it, and how they draw on prior knowledge and on the
simulation results, from which the agent learned as the run went on.

\begin{figure}[t]
  \centering
  \includegraphics[width=\linewidth]{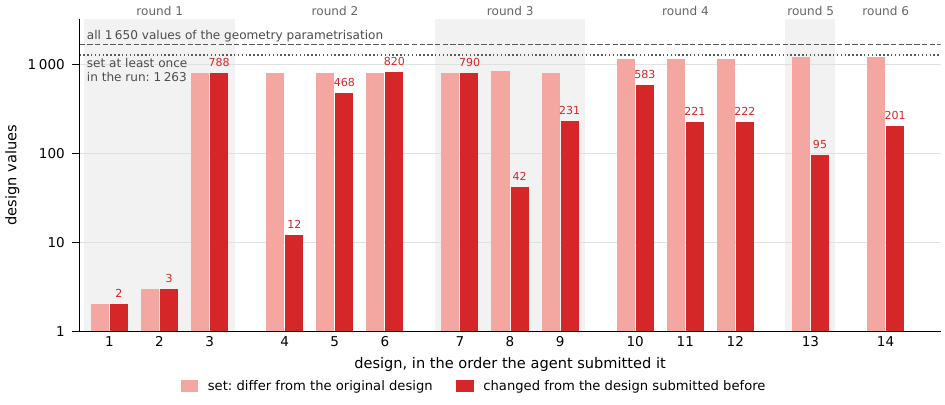}
  \caption{Number of design values the agent set in its clean run, out of the stent's about 1\,650, design by
  design. Light: the values in which a design differs from the original; dark, with their number: those in
  which it differs from the design submitted before it, within a round its sibling. Shaded: the six rounds.
  Dashed: all values of the geometry parametrisation; dotted: the values set at least once over the run,
  nearly all of them in edge profiles. Logarithmic scale.}
  \label{fig:valuesused}
\end{figure}

\paragraph{Extent of the changes.}
The parametrisation lets a design set about 1\,650 values, because each of the 60 crowns can be given its own
shape. From its third design on, the agent set about 800 of them, and from its tenth about 1\,200; over the
whole run it set 1\,263 of them at least once, nearly all in the edge profiles of the crowns
(\cref{fig:valuesused}). It did not set them one by one. Most designs gave the crowns only two or three
shapes, each shared by a whole family of crowns: one for the unlinked crowns, one for the linked crowns and
usually a third for the twelve crowns at the stent's ends. The many values come from writing each shape into
every crown of its family; written down one shape at a time, each as its differences from one already
written, such a design takes only 27 to 40 values. Three designs also varied single crowns: one in the middle
of the run varied the crowns' extension along the stent one by one, and the last two corrected single crowns,
the best design 20 of them, mostly with an extra bump of 0.01 to 0.02\,mm. The design validated in the
full-cycle model kept to two shapes, one for the unlinked and one for the linked crowns
(\cref{fig:pilotgeometry}). Choosing such a structure is what a person does before handing a classical
optimiser a handful of settings. The size of the changes followed the course of the run: the first round
tried three different concepts, the middle rounds refined one of them, mostly by changing hundreds of values
per design, and the last two rounds changed 95 and 201 values but moved no crown edge by more than
0.04\,mm, too little to change $Q$ beyond the noise of a rebuild.

\paragraph{Redesign rather than reshaping.}
We call a change a redesign when it alters a design's proportions, such as the width of the crown tips
against the struts or the length of the legs, and a reshaping, or polishing, when it moves the surface
locally to follow the current strain peaks within the given proportions, as a shape optimiser does
(\cref{sec:tosca}). Neither changes the layout of rings, crowns and links, which the parametrisation fixes.
The original design needs a redesign: its crown tips are almost a fifth narrower than its struts, so each
crown bends mostly at its tip, and it sits exactly on its volume and shortening limits. The agent's first
round compared three redesigns: much longer legs, which lowered $Q$ by 15\% but broke the lumen limit; far
less narrowed tips with shorter legs, which made $Q$ worse; and new edge profiles on every crown with the
eight links changed, which lowered $Q$ by 24\%. The middle rounds kept the third concept and lengthened the
legs, first by 13\% and then by 21\%. The best design therefore differs from the original in its
proportions: it adds 0.11\,mm of metal at the tip of every unlinked crown and 0.07\,mm on the inner edges of
the linked crowns, takes it from the shoulders, and has legs a fifth longer, while its metal volume stays
within 0.3\% of its limit and its lumen ends 0.003\,mm inside its limit. Only the last two rounds reshaped.
In the agent's own summary: ``The largest improvement came from spreading crown bending, not simply thinning
struts. Longer legs reduced strain but consumed recoil-diameter margin; excessive shoulder relief created new
strain peaks.''

\paragraph{Prior knowledge in the first move.}\label{sec:reading}
In the pilot sessions, a single call to the LLM with the original design's results proposed uniform
0.86\,mm struts without the narrowed tip, the one-shot fix from which Tosca also started (\cref{sec:tosca});
it lowered $Q$ by 7.5\%, as the LLM had predicted. Three variants of that call, one call each, show where the
idea came from. Without the table of per-crown results, the LLM proposed the same kind of change, about half
as far. With an anonymised prompt, the same numbers without the physics, it moved a link away from a
supported crown: it dodged an artefact of the supports instead of improving the design, with no gain and 66\%
more shortening. A repeat of the original call gave the same core change. The first move therefore came from
prior knowledge, and the results read without the physics led the LLM astray. Two of the three answers that
knew the part was a stent also lengthened it, which lowers its shortening as a percentage without any real
improvement; this is why the shortening limit is absolute (\cref{sec:objective}).

\paragraph{Use of the simulation results.}
Later in the run the agent corrected single crowns, guided by strain peaks it found in the element fields.
Whether such changes come from the simulation results is what the shuffled-results control tests, following
the misdirected-prompt ablation of \citet{wang2026calibrationagent}. We replayed the clean run up to the
results of a round and let the agent continue twice from the real results and twice from the same results
shuffled between crowns of the same kind, with each crown's element fields moved along with its numbers: the
results look plausible, but the most strained crowns appear in the wrong places. Each continuation ends with
its next design, which is built but not simulated, and the measures were fixed before any design was seen.
Started after the first round, the agent expected the strain where the shown results put it: of the 18 crowns
it predicted to be the most strained, three for each of its six designs, 14 were among the five that the
shuffled results showed as most strained and only one among the five that the simulation had actually found
most strained, against 12 of 18 with the real results. Its design changes in that round answered what the
shuffle leaves intact, such as the broken shortening limit, so, as stated in advance, they could not differ
between the variants. Started before the best design, where the clean run corrected single crowns, the
variants differ in the designs themselves: with the real results, both continuations aimed their changes at
the crowns that the simulation had found most strained, one widening crowns in line with their strain as the
clean run did, the other lengthening three of the five most strained; with the shuffled results, neither did.
The prompt, the geometry and the earlier rounds are the same in both variants, so these crown-level changes
come from the per-crown results, not from prior knowledge. Two smaller decisions also followed the shuffled
numbers to their new places: where the agent took margin on the stent's diameter and, in one continuation,
which linked crowns it reinforced. Its main reinforcement, however, did not follow the shuffled results to the
crowns shown as most strained; instead it changed shared settings, shorter legs and once wider struts, for
reasons the control does not show. With two continuations per variant and one round each, the control shows
a consistent pattern, not a statistic.

\paragraph{Learning within the run.}
The agent's predictions of its own results sharpened as the run went on. Its predicted $Q$ missed by 0.010 to
0.020 in the first round, two to five times the noise of a rebuild, and by at most 0.0035 from the fourth
round on, below that noise, though it stayed slightly optimistic: 12 of its 14 predictions were below the
outcome. Its predicted lumen missed by 0.03 to 0.045\,mm in the first round, several times the lumen's
tolerance, and by at most 0.005\,mm from the fourth, which let it end 0.003\,mm inside the lumen limit. Its
shortening predictions came down from misses of 0.09 to 0.13\,mm to about the scatter of a rebuild.

\subsection{Comparison with the industrial optimiser Tosca}\label{sec:tosca}

\begin{figure}[t]
  \centering
  \includegraphics[width=0.8\linewidth]{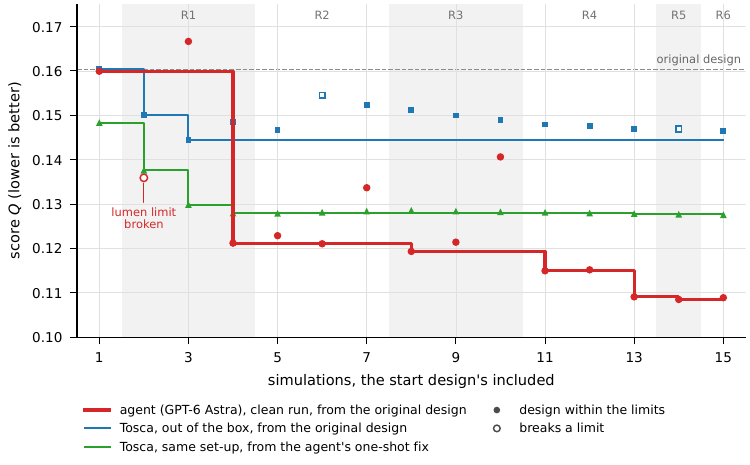}
  \caption{Optimisation progress on the stent: the score $Q$ of every design, the 90th percentile of its 60
  crown strains (lower is better), against simulations, all judged by the head-to-head reading of the limits
  (\cref{sec:objective}). Each marker is one simulated design, filled if it meets every limit and hollow if it
  breaks one; the line is the best score among the valid designs so far, as in \cref{fig:runs}. From the
  original design, fifteen simulations each: the
  agent's clean run, with its rounds R1--R6, and Tosca with its out-of-the-box set-up
  (\cref{sec:tosca-setup}). The agent's first simulation is the original design. Green: Tosca with the same set-up from the agent's one-shot fix, a design with
  uniform crowns, whose own simulation is its first; one run of fifteen simulations, capped after five to free
  the cluster and resumed by Tosca's own restart.}
  \label{fig:progress}
\end{figure}

\paragraph{Choice of comparison.}\label{sec:regime}
Classical optimisers cannot work on all of the stent's values within the budget (\cref{sec:intro,sec:related}). That leaves industrial shape optimisers, of which we ran
Tosca~\citep{tosca}. It moves all the surface nodes of the mesh, about 17\,600 on the stent, ten times the
agent's number of values, but it optimises what the solver computes on its node and element groups rather than
the judge's score and build rules, and its moves accumulate in one mesh rather than a CAD model, so it cannot
change proportions such as the leg length. Of its modules, shape optimisation is the one for the task; topology
optimisation would only remove metal inside the given struts or search a new layout from scratch.

\paragraph{Set-up.}\label{sec:tosca-setup}
We ran Tosca with its standard controller settings on the original design's mesh, with fifteen simulations and
the same judge as the agent, from two starts: the original design, and the agent's one-shot fix, a design with
uniform crowns that the LLM proposed in a single call from the original design's results
(\cref{sec:reading}). Tosca moves the nodes on the rings' cut faces within the tube surface; the links and
patches of 0.5\,mm around the links and supports stay fixed, and before each solve every node is returned to
its radius, so the wall keeps its thickness. Its target is the peak von Mises stress on the movable faces at
full expansion, with the metal volume held equal to the start's. Tosca does not see the lumen and shortening
limits; its designs are judged like the agent's, and those that break a limit do not count. These choices
follow Tosca's installed examples and are ours, not limits of the tool, and no run was tuned after its
results. Two set-ups closer to our score, with a plastic-strain target, the sensitivity-based method and the
lumen, shortening and volume limits, did not beat the original design in the simulations they ran, so we
report the out-of-the-box set-up, whose designs did.

\paragraph{Start from the original design.}
Tosca lowered $Q$, the 90th percentile of the crown strains, by 10\% within two updates, at its third
simulation (0.1444, valid only with the shortening tolerance), and then drifted (\cref{fig:progress}). Its own target kept falling, to 39\% below the start,
while the strain piled up in the fixed patches next to the reshaped regions, at the links and supports: the
most strained element reached an equivalent plastic strain of 0.47, about 2.7 times the end of the material data.
Beyond that end the material model holds the stress constant, so a stress target gets no signal, and the fixed
patches lie outside Tosca's target. $Q$, set by the most strained tenth of the crowns, stayed between 0.146 and
0.155 for the rest of the run.

\paragraph{Start from the agent's one-shot fix.}
Started from the agent's one-shot fix, whose crowns are uniform, Tosca lowered $Q$ by 14\% within four
simulations, steadily, and then held that level for its other ten without drifting: its best design, at its
fifteenth simulation, scored 0.1276 against 0.1279 at its fifth, a fourteenth of the noise. Every design was
valid under every reading of the limits (\cref{fig:progress}, green). From the original design's 0.1604, the
agent's fix and Tosca's polishing together reach 20\% lower, the agent's clean run alone 32\%. Tosca added metal
where the agent's edge profiles had added it in the pilot sessions, from a similar start with uniform crowns;
the two patterns of crown width correlate at 0.96. A third of its shape change was on the legs: it thinned
every leg, by 0.24\,mm on average from 0.86\,mm, and moved that metal onto the crown arcs. The agent's
geometry parametrisation cannot make that move, since a leg takes its crown's width; independent leg widths
were the first change the agent asked for at the end of its clean run (\cref{sec:extensions}). The run was capped at
five simulations to free the cluster, a decision taken before any of its designs was scored, and resumed to
fifteen by Tosca's own restart; we report it as one run. We report Tosca's designs as its own meshes give
them, but they are reshaped meshes, not rebuilt designs, so we checked its best design on a fresh mesh:
rebuilt as CAD from its own outline, each edge a radial wall as a laser cut makes it, and meshed afresh, it
scores 0.1295 against 0.1276, under half the noise apart, while its start, rebuilt the same way, moves only
from 0.1484 to 0.1476. Measured on fresh meshes at both ends, the polish is 12\% rather than 14\%.

\paragraph{Other measures.}
The objective is ours, so we also read each method's best valid design on four other measures, each against
the original design (\cref{tab:measures}). The three strain measures leave out the six crowns next to the
supports, as $Q_{\mathrm{ex}}$ does: the original's most strained element and crown sit there. The agent's
design leads on all of them. On the most strained element it is 32\% below the original, while Tosca's best
design from the original is 5\% above it. Tosca's run from the agent's one-shot fix ends 17 to 21\% below the
original design on every measure, the fix's own gain included.

\begin{table}[t]
  \centering
  \caption{Each method's best valid design on the stent on five measures: the change against the original
  design, solved in the method's own pipeline (lower is better), and the number of elements strained beyond
  the end of the material data. For Tosca's run from the agent's one-shot fix, the change includes the fix's own
  gain.}
  \label{tab:measures}
  \small
  \begin{tabular}{@{}lrrrrr@{}}
    \toprule
    Best valid design of & $Q$ & \shortstack[r]{Most strained\\element} & \shortstack[r]{Most strained\\crown} &
    \shortstack[r]{Median\\crown} & \shortstack[r]{Beyond\\the data} \\
    \midrule
    Agent, clean run & $-32\%$ & $-32\%$ & $-32\%$ & $-31\%$ & 0 \\
    Tosca, from the original design & $-10\%$ & $+5\%$ & $-3\%$ & $-11\%$ & 0 \\
    Tosca, from the agent's one-shot fix & $-20\%$ & $-17\%$ & $-21\%$ & $-20\%$ & 0 \\
    \bottomrule
  \end{tabular}
\end{table}

\paragraph{Interpretation and scope of the comparison.}
In the terms of \cref{sec:how}, Tosca's node moves are polishing, and polishing needs a sound design with room
to move. The original design offers neither: its crowns bend mostly at their narrowed tips; it sits exactly on
its volume and shortening limits, which are its own values; and its most strained spots lie at the end of the
material data, next to the fixed patches, where a stress target gets no signal. The agent's one-shot fix
changed these proportions, with uniform struts and no narrowed tips, and so left room on the volume and
shortening limits and no element beyond the end of the material data. From that start, Tosca's polishing
worked. The agent did both, redesigning and polishing, and chose which one a design needed.

The stent itself
does not favour Tosca: its objective and limits are measurements of the design, which the agent reads directly
and Tosca only through stand-ins; a design is a CAD model, rebuilt and meshed afresh for every proposal, while
Tosca's moves pile up in one mesh; the problem is strongly nonlinear, starts on its limits and rewards
coordinated changes of the proportions; and the agent sees the whole result, not a few responses and their
sensitivities.

We therefore do not claim that Tosca cannot optimise this stent: an expert rerunning it with a
changed set-up, such as a target closer to the judge's, gentler steps or a start with room on the limits, might
well succeed.

To rule out a fault in
how we run and judge Tosca, we also ran its own bundled example, a plate with a hole, an elastic part with a
local stress peak for which Tosca is the natural tool: run as shipped, it reproduced the vendor's result, and
with fifteen simulations instead of five it lowered the peak stress by 36\% on a common judge
(\cref{sec:ladder}, details in \cref{app:plate}).

\subsection{Comparison across problem sizes}\label{sec:ladder}
\Cref{tab:ladder} steps down in the number of values, from the stent to two problems that enter the same loop
(\cref{tab:interface}): Tosca's own plate with a hole, with up to 26 values, and a single crown with four. All
three have budgets of thirteen to fifteen simulations, the start's included, so the number of design values
per simulation falls from over a hundred to about two and to fewer than one.

\begin{table}[t]
  \centering
  \caption{Three problems with four, 26 and about 1\,650 values and similar budgets: the methods that can take part, and
  their results. Percentages: the best valid design's objective below the start design's, within the budget.
  Details in \cref{app:plate,app:crown}.}
  \label{tab:ladder}
  \small
  \begin{tabular}{@{}L{0.17\linewidth}L{0.25\linewidth}L{0.25\linewidth}L{0.25\linewidth}@{}}
    \toprule
    & Crown & Plate & Stent \\
    \midrule
    Design values & 4 & Up to 26 & About 1\,650 \\
    Proposed designs & 12 & 14 & 14 \\
    Ratio of design values to proposed designs & Below one & About two & Over a hundred \\
    \addlinespace
    Classical optimisers & On all four values: Bayesian optimisation 8.6--15.4\% (median 10.2\%), COBYLA
    0--5.8\%, random search nothing & Bayesian optimisation on the two semi-axes a person picks: 27\%, an
    ellipse & Not run as a comparison \\
    \addlinespace
    Shape optimiser (Tosca) & -- & As shipped, on its own mesh: 36\% & From the original: 10\%, then drifted;
    from the agent's fix: a further 14\% (12\% on a fresh mesh) \\
    \addlinespace
    LLMs & Complete sessions 10.1--22.3\%; the best of four LLMs 19--22\%; one LLM ahead of
    the median seed in two of three sessions & 27--47\%; 9 of 12 runs ahead of Tosca & 32\% (clean run), 26\%
    (pilot sessions) \\
    \addlinespace
    LLMs, runs & Six LLMs, 20 sessions & Four LLMs, three runs each & GPT-6 Astra, two runs \\
    \bottomrule
  \end{tabular}
\end{table}

\paragraph{Plate with a hole: Tosca's own example.}
Tosca's bundled shape example is a plate with a hole under two load cases, linear elastic, with the peak
stress at the hole's edge as the objective (\cref{app:plate}). Tosca ran it exactly as its vendor ships it,
with fifteen solves instead of five, and reached 36\% below the circle on a common judge. Bayesian optimisation,
given the hole's two semi-axes, stopped at an ellipse 27\% below the circle. The agents' geometry parametrisation adds up
to eight smooth bumps on the hole's edge, 26 values in all, more
than the fourteen designs of the budget. Four LLMs ran three times each: GPT-6 Astra (40 to 45\%),
Claude Fable~5.2 (46 to 47\%), and the open-weight Kimi~K3.1~Pro (33 to 40\%) and Mistral Large~4 (27 to
39\%). Nine of the twelve runs beat Tosca, all of Astra's and Fable's among them, and the four best runs were
the four that placed all eight bumps freely. Given Tosca's own coarse model instead of the judge's, seven of
ten agent runs still beat it. On its home ground Tosca is a real competitor, which is what makes the agents'
lead count.

\paragraph{Single crown: fewer values than solves.}
On a single crown with four variables and twelve simulations after the start (\cref{app:crown,fig:crownmodel}), random search found no
feasible improvement and COBYLA at most 5.8\%, but Bayesian optimisation reached 8.6 to 15.4\% over five
seeds. The best sessions of four of six LLMs lowered the peak strain by 19 to 22\%, and every
session that used its full budget was level with the median seed or ahead of it; but only one LLM did
better than the median seed in two of its three sessions, the criterion fixed before the baselines ran.

\paragraph{Interpretation of the ladder.}
With fewer values than solves, a good classical optimiser can search the whole design space, and the LLMs'
lead becomes a matter of the run. Once the values outnumber the solves, the classical optimisers need a
person to shrink the problem and stop where the shrunken space ends, and only the shape optimiser keeps up, on
its home ground. At over a hundred values per solve, only the shape optimiser can take part, and from the
original stent it could not redesign. The problems differ in more than their size, so the ladder is not a
scaling law; it shows where each family can take part, and how it did.

\section{Discussion}\label{sec:discussion}

\paragraph{Sources of the agent's advantage.}
The agent brings what the optimisers lack: physical knowledge for its first move, the full fields of every
solve, and the ability to choose a structure and a scale for its moves. Its first idea came from prior
knowledge; its later, crown-level moves came from the results it read, as the shuffled-results control shows,
with predictions that sharpened round by round. Unlike LLM calibration, where an
early lead faded within about fifteen evaluations~\citep{wang2026calibrationagent}, the agent kept improving
over its fifteen solves. We attribute this to the number of values: with many values there is always another
informed move to make, and the agent can choose how large it should be.

\paragraph{Position among automated methods.}
To our knowledge, the agent's loop is the only automated one that meets all three conditions; every family of
automated optimisers gives up one of them (\cref{fig:conditions}). The automated method that would also meet
all three, a gradient method on the CAD values with adjoint gradients of the judge's own measures through
plasticity and contact (\cref{sec:highdim}), does not exist off the shelf. Grey-box Bayesian
optimisation~\citep{astudillo2019composite}, which models per-crown outputs rather than the score alone, is the
closest machine-learning idea to reading the results, but it still scales with the number of design values.

\paragraph{Classical optimisers on the agent's structure.}
The agent did not turn 1\,650 knobs one by one: from its third design on it set about 800 to 1\,200 values, in
most designs as two or three crown shapes that take only 27 to 40 values to write down, a structure a
classical optimiser could be given. But choosing that structure from the results of the original design is part of what the agent did; a
classical optimiser needs a person to hand it one, and the agent changed its structure between designs,
working on single crowns in three of them. Where every method had the
same four numbers, Bayesian optimisation was a real competitor (\cref{app:crown}).

\paragraph{Division of labour with shape optimisers.}
The agent redesigns and polishes, and decides which one a design needs; Tosca's shape optimisation polishes,
and polished the agent's first redesign well. A natural division of labour follows, as in hybrids of LLMs and
optimisers~\citep{ferreira2026centaur,wang2026calibrationagent}: the agent finds where to act and a shape
optimiser tunes how much.

\paragraph{Remaining human input.}\label{sec:extensions}
The set-up is kept out of the agent's hands on purpose. When LLMs had to write a simulation set-up
themselves, for a bar whose optimum is known, five of six produced a working study in most of their
sessions, but the weakest accepted set-ups that ran and measured something else, and reported success
(\cref{app:crown}). A new problem needs its four parts, of which the geometry parametrisation is the most
work. The agent helps here, because at the end
of each session it lists the changes it wanted but could not express. In the pilot sessions, after
crown-by-crown editing, it asked for ``Localized, smoothly blended shoulder reinforcement'' and
``Noncircular/variable-curvature inner crown profiles'', which the edge profiles of \cref{sec:stent-language}
then provided; at the end of the clean run it asked for ``Independent leg-width/curvature control,
noncircular crown centerlines, curved link centerlines or different attachment topology, and
through-thickness edge fillets''. A parametrisation can thus be extended where a problem needs it, between
runs and never within one.

\paragraph{Limitations.}
The main limitation is that the agent changes the CAD model only through a geometry parametrisation written for
this geometry. The parametrisation decides what the agent can change, and so bounds what it can find, and every
new part needs its own (\cref{sec:stent-language}). The loop also takes time, so only one LLM,
GPT-6 Astra, ran on the stent, and only twice: the clean run took 35 hours from its first simulation to its
last (\cref{app:stent-model}). How much the result varies from run to run is therefore open. Of the three problems considered, only the stent is a nonlinear
three-dimensional part; the crown is a planar model with four values, and the plate is linear elastic. The
original stent derives from the Open Stent Design~\citep{bonsignore_openstent}, an openly published research
design rather than a state-of-the-art commercial one, so whether an agent can improve a state-of-the-art design
remains open; as a test of the capability itself, a standard design serves. The shuffled-results control has
two continuations per variant and one round each, and with shuffled results the agent's main reinforcement did
not move to the crowns shown as most strained (\cref{sec:reading}). Finally, our finite-element model covers
the expansion alone; the full-cycle check supports its gains for one design but cannot establish them in
general (\cref{sec:fullcycle}).

\section{Conclusion}\label{sec:conclusion}
We asked whether a general-purpose LLM, working in a loop with CAD and finite-element simulation, can redesign a
nonlinear three-dimensional part within a budget of fifteen simulations. On a magnesium stent it did: through
about 1\,650 values of the stent's CAD model, the agent lowered the strain objective by a third for US\$19 of
LLM calls, a gain an order of magnitude larger than the noise of a rebuild. A design from its pilot sessions
held up in an independent, higher-fidelity model of the full deployment with crimping and damage, which the
agent never saw: it was 14\% less strained than the original, and its most damaged spot began to fail later
and carried 63\% less damage at full expansion. To our knowledge, this is the first stent designed by an LLM
agent, and the agent's loop is the only automated one that combines freedom to redesign with a high parameter
count and a small simulation budget.

The agent did not search blindly. It chose its own structure, giving whole families of crowns the same shape
and correcting single crowns only at the end, and the scale of each move, with different concepts first and
polishing last. Its first move came from prior knowledge; its later, crown-level moves came from the results it
read, as the shuffled-results control shows, and its predictions sharpened as the run went on. The comparisons
place this capability. Tosca, the industrial shape optimiser that can work at this level of detail and budget,
polished well but could not redesign: from the original design it gained a tenth and then stopped improving,
while from a design the agent had already redesigned it reached 20\% below the original. On Tosca's own bundled
example, run as shipped, agents built on four LLMs beat it in 9 of 12 runs, and on a single crown with fewer
values than simulations a good Bayesian optimiser caught up with them. Agents thus lead where the design values
outnumber the simulations, and they combine naturally with shape optimisers: the agent redesigns, the optimiser
polishes.

Two limits remain. A person still writes the simulation set-up, because weaker LLMs accept wrong set-ups and
report success, and every new part needs a geometry parametrisation through which the agent can change its CAD
model; the agent eases the latter by naming the changes it could not express. Whether the capability carries
over to other parts, to other LLMs on the stent and to designs that are already state of the art remains to be
tested. Engineering CAD, with its hundreds to thousands of values, may be where reading the simulation pays
most.

\section*{Acknowledgements}
We thank Dirk Steglich (Helmholtz-Zentrum Hereon) for the stent that this work starts from, and for running
the agent's design and the original design through his full-cycle model of crimping and
expansion~\citep{steglich2026loadreversals} and sharing its results. We gratefully acknowledge the computing
time on Strand, the HPC cluster of Helmholtz-Zentrum Hereon, operated by its department Logistics and
Organization Coastal Research. This work was supported by the Digital
Twins of Complex Systems (DToCS) Platform at Helmholtz-Zentrum Hereon, which is jointly funded by the German
Federal Government and the state governments of Schleswig-Holstein, Hamburg, Lower Saxony, and Brandenburg.

\section*{Code and data availability}
The geometry parametrisation and its checks, the agent's tools and prompts, the scripts that score and plot the
results, and the session records are available from the authors on reasonable request. The simulations need
the commercial codes Abaqus and Tosca.

\appendix

\section{Protocols, prompts and tools}\label{app:protocols}
\Cref{tab:interface} lists, for each of the three problems, what the agent receives and writes, how its
designs are simulated and judged, and its budget. This appendix gives the settings that each run's protocol
fixed, the system prompt, opening message and tools of the clean run on the stent, a round of its transcript, how
the prompts and tools of the other runs differ, and what the records allow to be reproduced.

\paragraph{Protocols.}
The record of each run states what was fixed before its first LLM call: the LLM and its settings, the opening
message, the tools, the budget, the cost caps, and the questions the run is to answer, with how each answer
will be read. \Cref{tab:protocols} lists the settings of every run of the agent in this paper. All LLMs
were called through OpenRouter. On the stent and the plate, a session ends before any call whose worst case, its
whole prompt and the longest reply allowed, could exceed the cap per call or what is left of the session's cap;
a turn is one call. Two thirds of the clean run's cost went on the six calls that followed its rounds of
simulations: the prompt cache does not last through hours of simulation, so each of these calls paid in full
for reading the whole conversation. Changes made after a run had started are recorded with their dates and
reasons. On the
plate, during the first run of Claude Fable~5.2, the cap per call was raised from US\$12 to US\$25 and the turn
limit from 150 to 300, for that session and every later one: the run needed about twelve turns per design and
would otherwise have been stopped before its fourteenth design, and the budget compared between methods is the
designs, not the turns. Claude Fable's second run first stopped at its third call, which the provider refused as
overloaded on every retry; that attempt is kept in the records, and the run was repeated from the start. The
crown's amendments are described in \cref{app:crown}.

\begin{table}[tbp]
  \centering
  \caption{The agent's runs and the settings of their protocols; changes during a run are given in the text.
  Reasoning effort as sent to OpenRouter; the crown's LLMs ran at temperature 0.2 with their default reasoning.
  Budget: simulations of the agent's designs, beyond the start design's own; checks: free design checks. Caps
  and costs in US\$ of LLM calls, per session unless stated; a turn is one LLM call. Costs as billed, for the
  runs on the stent and the crown, all sessions of a row together; the single call's include its three variants,
  and the crown's its set-up sessions, which were billed together.}
  \label{tab:protocols}
  \small
  \begin{tabular}{@{}L{0.16\linewidth}L{0.2\linewidth}L{0.27\linewidth}L{0.19\linewidth}r@{}}
    \toprule
    Run & LLM & Budget and tools & Caps & Cost \\
    \midrule
    \multicolumn{5}{@{}l}{\textit{Stent: pilot sessions (\cref{sec:headline})}} \\
    Single call & GPT-6 Astra, high & One reply, no tools, whose design is simulated; also three one-call
    variants (\cref{sec:reading}) & 6 per call, 15 in all & 2.86 \\
    Four shared settings & GPT-6 Astra, high & Three proposals, one call each, each after the results of the
    one before & 6 per call, 15 in all & 1.69 \\
    Crown by crown & GPT-6 Astra, max & Three simulations, one design at a time; 40 checks; tools as in
    \cref{tab:tools} & 150; 12 per call; 150 turns & 7.31 \\
    Edge profiles & GPT-6 Astra, max & Nine simulations in rounds of up to three; 60 checks & 150; 12 per call;
    150 turns & 17.47 \\
    \addlinespace
    \multicolumn{5}{@{}l}{\textit{Stent: the finished loop}} \\
    Clean run & GPT-6 Astra, max & 14 simulations in rounds of up to three; 60 checks & 150; 12 per call;
    150 turns & 19.44 \\
    Shuffled-results control & GPT-6 Astra, max & Eight continuations of the clean run, each ending at its next
    round, whose designs are built but not simulated & 25 per continuation, 60 per arm; 20 turns & 14.99 \\
    \addlinespace
    \multicolumn{5}{@{}l}{\textit{Plate (\cref{app:plate})}} \\
    On the judge's fine model & GPT-6 Astra, Claude Fable~5.2, Kimi~K3.1~Pro, Mistral Large~4, max; three
    runs each & 14 simulations in rounds of up to three; 60 checks & 200 to 260; 25 per call; 300 turns & -- \\
    On Tosca's coarse model & The same, but Claude Fable~5.2 one run & The same & 200; 25 per call; 300 turns
    & -- \\
    \addlinespace
    \multicolumn{5}{@{}l}{\textit{Crown (\cref{app:crown})}} \\
    Crown sessions & Six LLMs, 20 sessions & 12 simulations in rounds of up to three, after a critic's review;
    no sandbox, no CAD & 0.25 to 30 by LLM and round; 200 or 1\,000 steps; 150 or 240 minutes & 98.61 \\
    Set-up sessions & Six LLMs, 20 sessions & The LLM writes the study, then 12 simulations in rounds of up to
    three & 0.50 to 30 by LLM and round; 200 or 1\,000 steps; 150 minutes & \\
    \bottomrule
  \end{tabular}
\end{table}

\paragraph{System prompt and opening message.}
Every session with tools on the stent and the plate had the same system prompt: ``You are working on an
engineering design task with the tools provided. Use them as you see fit and give your final answer with the
finish tool.'' The
opening message states the problem in seven sections: how the design is generated, how a design file changes
it, the simulation set-up, the start design with its results, the objective and the limits, the tools and the
budget, and what to report with each design. It states facts only, with no hypothesis about the results and, in
the clean run, nothing from the pilot sessions. The clean run's opening message follows as it was sent, except
that its table of the 60 crowns is cut to its first four rows. It gives the limits of the main reading
(\cref{sec:objective}): the agent saw the verdicts of that reading only, not those of the strict or the
head-to-head reading. The start design D0 in the message is the original design as the
agent's pipeline rebuilt it, which scored 0.1599 rather than the 0.1604 quoted elsewhere, a gap well within
the noise of a rebuild (\cref{sec:noise}).

\Needspace{12\baselineskip}
\begin{Verbatim}
# A magnesium stent: edit the design crown by crown, with tools and 14 simulations

You get a stent design problem, its original design D0 and D0's recorded deployment. You can change the
shared settings, single crowns and single links, including the shape of each crown's edges, and you work
with tools: a Python sandbox with the full recorded fields of every simulated design, a free design check,
and a budget of 14 deployment simulations, run in rounds of up to three at once. Section 6
describes the tools.

## 1. The design generator

A balloon-expandable stent is laser-cut from a magnesium-alloy tube: outer radius
R = 7.995125 mm, wall thickness t = 0.59285 mm (inner radius 7.402275 mm).
The cut pattern is drawn on the outer tube surface, unrolled (axial coordinate z, arc length s),
and wrapped onto the tube with radial walls (every wall points at the tube axis).

- 5 rings. Each ring is a serpentine of 12 semicircular crowns (6 at each ring end),
  joined by straight axial legs. Crowns alternate between the two ring ends every
  360/12 = 30 degrees around the tube.
- Crown period along the surface: half = 2 pi R / 12 = 4.186238 mm. For strut width w,
  the crown inner radius is r_in = (half - w)/2 and the outer radius r_out = (half + w)/2.
- crown_offset d: each crown's outer arc is shifted by d towards the crown opening, so the crown
  tip is narrower than its sides (tip width w - d).
- leg_length L: straight leg between consecutive crowns, width w.
- Ring width (axial) = L - d + 2 r_out. The ring_pitch p sets the distance between ring centres;
  the axial gap between rings is p minus the ring width.
- Neighbouring rings are shifted by half a crown period, so crown tips face each other across
  each gap. Hourglass links join facing crown tips: link_crowns = ((2, 5), (0, 3)) lists the
  link-carrying crowns of the left ring per gap, cycled over the gaps; this gives two links per
  gap, 180 degrees apart, rotated by 90 degrees from one gap to the next. Link waist width
  wl = 0.74315 mm with a straight waist flat f = 0.280963 mm; each link flank is an arc
  tangent to the facing crown's outer arc and to the waist, with radius
  r_link = (dx^2 + (wl/2)^2 - r_out^2) / (2 (r_out - wl/2)), where dx = p/2 - (L/2 - d/2) - f/2.
- The minus-z crown tips of ring 0 are fixed at z = -27.792043 mm; the first ring centre is at
  z = -27.792043 + r_out + L/2 - d/2 and ring k at that plus k p. As-cut length = 4 p + ring width.
- Metal volume V = A (R^2 - R_in^2)/(2R) = 0.570870 * A for the unrolled pattern area A.

## 2. Editing the design: shared settings, single crowns and single links

A design is a JSON file (schema "astra-direct-002/design-1"): shared settings plus optional entries for single crowns
and links. A crown or link without an entry, and any field left out, takes the shared value. D0, the
original design, is
{"schema": "astra-direct-002/design-1", "settings": {"strut_width_mm": 0.96314, "crown_offset_mm": 0.178238, "leg_length_mm": 0.773662, "ring_pitch_mm": 8.2498}, "crowns": [], "links": []}

- Crowns are named (ring, theta_deg) and links (gap, theta_deg) exactly as in the result tables below
  (gap g joins rings g and g+1; a link's theta is that of its crowns).
- Crown fields: width_mm (strut width of the crown and its shoulders; the legs taper between
  neighbouring crowns of different width), tip_narrowing_mm (the crown's own offset d: tip width =
  width - d; negative values widen the tip; shoulders and legs keep the crown's width), extension_mm
  (the whole crown moved outwards along the axis, lengthening its two legs; negative shortens them),
  and profile (below).
- profile: a list of smooth bumps on the crown's edges, each {"edge": "outer" or "inner",
  "center_deg": angle from the crown's apex (+ towards larger theta), "half_width_deg": reach to either
  side, "amount_mm": material added at the centre (negative removes)}. The outer edge (on the crown's
  tip side) moves out and the inner edge (on its opening side) moves in by amount * (1 - t^2)^3,
  t = (angle - center)/half width, so the edge returns smoothly to the arc. Bumps must end within
  +-85 degrees of the apex and 2 degrees before the legs; on a linked crown's outer edge they must
  stay 5 degrees clear of the link attachment (about +-28 degrees in D0). Up to 6 bumps per crown;
  center_deg -80 to 80, half_width_deg 5 to 60, amount_mm -0.3 to 0.3; the local width along the crown must
  stay >= 0.57 mm.
- Link fields: waist_width_mm, waist_flat_mm (a longer flat gives smaller flank radii, i.e. a narrower
  attachment at the crowns) and shift_mm (the link's axis moved along the circumference, + towards
  larger theta). Each flank is fitted to its own crown, so the two crowns of a link may differ.
- Legs are common tangents of neighbouring crown arcs and every outline stays tangent-continuous.
  The legs are short, so neighbouring crowns can differ in width only by about 0.14 mm at L = 0.77 mm
  before a leg vanishes; the check reports it.
- Shared settings: strut_width_mm 0.76 to 1.1, crown_offset_mm 0.0 to 0.4, leg_length_mm 0.4 to 1.4, ring_pitch_mm 7.6 to 8.75, link_waist_width_mm 0.57 to 1.2, link_waist_flat_mm 0.0 to 2.0;
  ring_pitch_mm may be one value or four, one per gap.
- Per crown: width_mm 0.76 to 1.2, tip_narrowing_mm -0.3 to 0.63, extension_mm -0.3 to 0.4.
  Per link: waist_width_mm 0.57 to 1.2, waist_flat_mm 0.0 to 3.0, shift_mm -0.5 to 0.5.
- Fixed: the tube, 5 rings, 12 crowns per ring, which crowns carry links, the z of ring 0's minus-z
  crown tips in the shared construction. Rules: crown tip width >= 0.57 mm, every link flank radius
  >= 0.3 mm, every leg's straight part >= 0.2 mm, facing unlinked crown tips >= 0.5 mm
  apart, link attachments inside the crown arcs (>= 3 degrees from the legs), the profile rules
  above, as-cut length <= 41.0 mm, volume <= 229.193 mm^3.
- /astra_lib/local_stent.py builds the outline from a design and runs these checks; read it for the
  exact construction (resolve, check, measures, crown_table, plot).

## 3. The simulation (set-up, the same for every design)

- Abaqus/Standard, quasi-static, large deformation. C3D8 hexahedra swept through the wall
  (3 layers, graded 1:2:1), in-plane element size 0.19 mm (about 37,000 elements).
- Material: elastic-plastic, von Mises yield with isotropic hardening, E = 43 GPa, nu = 0.3.
  The hardening curve is calibrated up to an equivalent plastic strain (PEEQ) of 0.1766; above
  that the model is outside its calibration (it is not a fracture criterion).
- Analysis frame: stent axis = z, angle theta = atan2(y, x) in degrees.
- Balloon: a cylindrical surface, radius 3.80 mm, length 42 mm, centred on the stent; its nodes
  are moved radially. Steps: Approach (balloon brought to 0.05 mm from the stent's inner surface),
  Expansion (balloon radius increased by 7.0 mm, i.e. balloon diameter 21.6 mm), Recoil (balloon
  returned to its initial radius; the stent springs back freely).
- Contact: hard normal contact; friction 0.05 between balloon and stent; frictionless stent
  self-contact.
- Rigid-body motion is removed by three single nodes on the outer surface of ring 2, near target
  angles 0, 120 and 240 degrees on the accepted node plane nearest to 0.09 mm below the ring
  centre. Their positions depend on each design's mesh (listed below). At each, the tangential
  and axial displacements are fixed, the radial one is free.
- The swept mesh needs a cut through the tube along the plane y = 0 (theta = 0 and 180 degrees);
  it passes through the crown apices at theta 0 and 180 in every ring. A few slightly distorted
  elements lie next to this cut.

## 4. The original design D0

D0 is the design of section 2: shared settings only, no crown or link entries. It was simulated with
this set-up; its design file and recorded fields are in /data/designs/D0.

Geometry of D0:
id,origin,strut_width_mm,crown_offset_mm,leg_length_mm,ring_pitch_mm,tip_width_mm,ring_width_mm,ring_gap_mm,link_flank_radius_mm,length_as_cut_mm,volume_mm3
D0,original design,0.963140,0.178238,0.773662,8.249800,0.7849,5.7448,2.5050,1.6115,38.744,229.193

Results (after recoil). Crown strain = mean of the 5 highest element PEEQ values in a crown (a crown
here is the arc plus the adjacent halves of its two legs); Q = 90th percentile of the crown strain over
all 60 crowns; shortening = as-cut length minus length after recoil. As-cut inner diameter 14.805 mm for
every design. feasible: under the limits of section 5.
id,Q,median_crown_strain,max_crown_strain,max_element_peeq,elements_above_0.1766,inner_D_after_recoil_mm,recoil_pct,shortening_mm,feasible
D0,0.1599,0.1511,0.1802,0.1857,3,21.2591,2.090,1.3294,yes

Support nodes:
id,support_theta_deg,support_z_mm
D0,348.45 / 131.55 / 228.45,-8.509 / -8.509 / -8.509

Per crown: crown strain, max and mean element PEEQ, and tip_inward_mm = axial motion of the crown tip
towards its own ring centre after recoil. linked = a link is attached at the crown tip.
ring,theta_deg,tip_end,linked,crown_strain_D0,D0_max_element_peeq,D0_mean_peeq,D0_tip_inward_mm
0,0,-z,no,0.1454,0.1470,0.0263,+1.031
0,30,+z,yes,0.1449,0.1484,0.0246,-0.022
0,60,-z,no,0.1450,0.1474,0.0261,+1.048
0,90,+z,no,0.1567,0.1600,0.0290,+0.723
[... 56 further rows, one per crown ...]

D0 in more detail. Inner mean diameter 21.713 mm at the end of expansion; length
38.742 mm as cut, 36.736 mm at the end of expansion, 37.413 mm after recoil; max von Mises
stress 247.5 MPa at the end of expansion, 131.4 MPa (residual) after recoil. PEEQ does not
change during recoil.

Per link (D0):
gap_between_rings,theta_deg,max_element_peeq
0-1,30,0.0160
0-1,210,0.0441
1-2,120,0.0152
1-2,300,0.0504
2-3,30,0.0154
2-3,210,0.0236
3-4,120,0.0195
3-4,300,0.0415

Per ring (D0):
ring,centre_z_mm,inner_D_expansion_end_mm,inner_D_after_recoil_mm,crown_tip_to_tip_length_as_cut_mm,after_recoil_mm
0,-24.920,21.713,21.258,5.598,4.147
1,-16.670,21.710,21.259,5.615,4.165
2,-8.420,21.706,21.254,5.618,4.182
3,-0.170,21.710,21.261,5.615,4.164
4,8.080,21.713,21.260,5.582,4.118

The full recorded fields of D0 (every node and element, every exported frame) are in the sandbox under
/data/designs (see /data/designs/README.md); each design you simulate is added there.

## 5. Objective and rules

Objective: the lowest Q of a feasible design. D0 (Q 0.1599) is the design to beat; the best feasible
design of this session counts.

Constraints (feasible = all met):
- inner mean diameter after recoil >= 21.249 mm;
- shortening <= 1.3605 mm;
- volume <= 229.193 mm^3 (D0's volume); as-cut length <= 41.0 mm;
- the design rules of section 2.

A design that cannot be built, meshed or solved counts as Q = 0.25 and infeasible.

## 6. Tools and budget

- python: runs one Python program in an isolated sandbox (numpy, scipy, matplotlib, shapely; no network;
  up to 10 minutes per run). /data and /astra_lib are read-only; /work is yours and persists between runs.
- view_image: shows you a PNG or JPEG you saved under /work.
- check_design: the free checks of section 2 plus the CAD build of a design file under /work; it
  reports the volume, length, link flank radii, profiles and what changed relative to D0. Up to 60 checks.
- submit_designs: simulates one to 3 designs at once (about 2.5 to 3 hours per round; the conversation
  waits) and returns all their results; their full fields then appear under /data/designs. Give your
  predictions and reasons with each design. 14 simulations in total; every design must pass check_design
  first and must not repeat a simulated one.
- finish: ends the session with your best design and a summary.

Between turns you keep your messages, tool calls and their results, and short summaries of your
reasoning, not your full reasoning; keep notes under /work if you want to.

## 7. What to report

With every design you submit, predict Q, the inner mean diameter after recoil, the shortening, the max
crown strain and the three hottest crowns, and give your reasons. When you are done, at the latest after
the results of your last simulations, call finish: name your best design, summarise what you learned,
and list any change you wanted to make but could not express with these controls.
\end{Verbatim}

\paragraph{Tools.}
\Cref{tab:tools} lists the clean run's five tools as the LLM received them. With every call, the LLM gets back
its earlier messages, tool calls and results, and summaries of its earlier reasoning, as its opening message
says. The sandbox is an isolated Linux system without network access: \texttt{/data} holds the fields of every
design simulated so far, \texttt{/astra\_lib} the design code and a small library to read the fields, and
\texttt{/work} whatever the agent writes. A tool's result is cut in the middle beyond 30\,000 characters, the
result of a round beyond 60\,000. At most three times in a session, a reply without a tool call is answered with
``Please continue with the tools, or call finish when you are done.'' The clean run needed none.

\begin{table}[tbp]
  \centering
  \caption{The clean run's tools, with the descriptions the LLM received, verbatim. The plate's sessions had the
  same tools, with the descriptions and predictions of their problem.}
  \label{tab:tools}
  \small
  \begin{tabular}{@{}L{0.17\linewidth}L{0.44\linewidth}L{0.33\linewidth}@{}}
    \toprule
    Tool & Description & Arguments \\
    \midrule
    \texttt{python} & ``Run one Python program in the sandbox and return what it prints. /data and /astra\_lib
    are read-only; /work is writable and persists between runs. No network; up to 10 minutes per run.'' &
    \texttt{code}: ``Python source'' \\
    \addlinespace
    \texttt{view\_image} & ``Show a PNG or JPEG file saved under /work; it appears in the next message.'' &
    \texttt{path}: ``e.g. /work/outline.png'' \\
    \addlinespace
    \texttt{check\_design} & ``Free checks and CAD build of a design file under /work (schema
    astra-direct-002/design-1). Reports pass or fail, the failures, volume, length, link flank radii, profiles,
    what changed relative to D0 and whether it was already simulated. Up to 60 checks.'' & \texttt{path}: ``e.g.
    /work/d8.json'' \\
    \addlinespace
    \texttt{submit\_designs} & ``Simulate one to 3 designs at once, in parallel (about 2.5 to 3 hours per round;
    the conversation waits), and return all their results. Each must pass check\_design and must not repeat a
    simulated design. 14 simulations in total.'' & \texttt{designs}: one to three, each with \texttt{path}
    (``design file under /work''), \texttt{reasons} (``why this design'') and \texttt{predictions} of
    \texttt{Q}, \texttt{inner\_d\_after\_recoil\_mm}, \texttt{shortening\_mm}, \texttt{max\_crown\_strain} and
    \texttt{hottest\_crowns} (``the three hottest crowns, hottest first, as `ring@theta'\,'') \\
    \addlinespace
    \texttt{finish} & ``End the session.'' & \texttt{best\_design} (``id of your best design, as named in the
    results''), \texttt{summary} and \texttt{wanted\_but\_not\_expressible} (``changes you wanted but could not
    express with these controls'') \\
    \bottomrule
  \end{tabular}
\end{table}

\paragraph{A round in the transcript.}
The clean run's transcript holds the system prompt, the opening message, 30 replies of the LLM and the results
of its 38 tool calls: 16 Python programs, one image, 14 checks, six rounds and the finish. In its eighth reply,
after five Python programs, one image and three checks, the LLM opened its first round with the call below,
whose three designs are the three concepts of \cref{sec:how}. The tool's reply follows, cut like the opening
message. Its verdicts are those of the main reading; under the head-to-head reading, only the first design,
which breaks the lumen limit, is invalid. The run ended with the LLM's call of \texttt{finish}, the last
listing.

\begin{Verbatim}
{
 "designs": [
  {
   "path": "/work/longlegs.json",
   "predictions": {
    "Q": 0.146,
    "inner_d_after_recoil_mm": 21.222,
    "shortening_mm": 1.24,
    "max_crown_strain": 0.168,
    "hottest_crowns": ["2@210", "3@150", "2@180"]
   },
   "reasons": "Increase straight-leg length to 1.15 mm to reduce the crown rotation required by hoop expansion; reduce shared width to 0.922944 mm to hold volume at 228.90 mm3. This probes the strain/recoil tradeoff, with anticipated diameter risk."
  },
  {
   "path": "/work/less_taper.json",
   "predictions": {
    "Q": 0.148,
    "inner_d_after_recoil_mm": 21.251,
    "shortening_mm": 1.3,
    "max_crown_strain": 0.173,
    "hottest_crowns": ["2@210", "1@210", "3@150"]
   },
   "reasons": "Reduce the tip narrowing from 0.178238 to 0.04 mm, making the tip relatively stronger and shoulders thinner at fixed volume. Shorten legs to 0.58 mm to offset extra compliance and test whether a broader bending region lowers the apex peak while maintaining recoil diameter."
  },
  {
   "path": "/work/redistribute.json",
   "predictions": {
    "Q": 0.141,
    "inner_d_after_recoil_mm": 21.239,
    "shortening_mm": 1.33,
    "max_crown_strain": 0.158,
    "hottest_crowns": ["2@210", "3@150", "1@210"]
   },
   "reasons": "Use the D0 element fields to redistribute material: unlinked crowns receive a broad outer apex addition and outer shoulder relief; linked crowns receive small inner additions at +/-30 degrees and outer shoulder relief. Longer link waist flats narrow the attachments. This directly tests spreading plastic curvature away from the measured hot zones."
  }
 ]
}
\end{Verbatim}

\begin{Verbatim}
Round of D1 (d8e49580ef84), D2 (9e89cacbac8f), D3 (1675e130361e):

Results (after recoil):
id,Q,median_crown_strain,max_crown_strain,max_element_peeq,elements_above_0.1766,inner_D_after_recoil_mm,recoil_pct,shortening_mm,feasible
D1,0.1359,0.1295,0.1465,0.1520,0,21.1847,2.433,1.1460,no
D2,0.1667,0.1589,0.1725,0.1764,0,21.2958,1.934,1.4296,no
D3,0.1212,0.1153,0.1412,0.1465,0,21.2719,2.143,1.4397,no
D1: infeasible: inner diameter after recoil 21.1847 mm < 21.2491 mm. Predicted against actual: Q 0.146 / 0.1359; inner D after recoil 21.222 / 21.1847 mm; shortening 1.24 / 1.1460 mm; max crown strain 0.168 / 0.1465; hottest crowns 2@210, 3@150, 2@180 / 2@30, 3@150, 2@60.
D2: infeasible: shortening 1.4296 mm > 1.3605 mm. Predicted against actual: Q 0.148 / 0.1667; inner D after recoil 21.251 / 21.2958 mm; shortening 1.3 / 1.4296 mm; max crown strain 0.173 / 0.1725; hottest crowns 2@210, 1@210, 3@150 / 2@330, 2@270, 2@300.
D3: infeasible: shortening 1.4397 mm > 1.3605 mm. Predicted against actual: Q 0.141 / 0.1212; inner D after recoil 21.239 / 21.2719 mm; shortening 1.33 / 1.4397 mm; max crown strain 0.158 / 0.1412; hottest crowns 2@210, 3@150, 1@210 / 2@210, 2@180, 3@180.

Support nodes:
id,support_theta_deg,support_z_mm
D1,11.69 / 108.31 / 251.69,-8.341 / -8.341 / -8.341
D2,18.24 / 134.24 / 221.76,-8.554 / -8.546 / -8.554
D3,348.45 / 131.55 / 228.45,-8.509 / -8.509 / -8.509

Per crown (crown strain of D0 and of this round; the rest for D3):
ring,theta_deg,tip_end,linked,crown_strain_D0,crown_strain_D1,crown_strain_D2,crown_strain_D3,D3_max_element_peeq,D3_mean_peeq,D3_tip_inward_mm
0,0,-z,no,0.1454,0.1245,0.1549,0.1110,0.1132,0.0276,+1.097
0,30,+z,yes,0.1449,0.1243,0.1517,0.1138,0.1153,0.0259,-0.097
0,60,-z,no,0.1450,0.1226,0.1498,0.1081,0.1096,0.0286,+1.105
0,90,+z,no,0.1567,0.1229,0.1558,0.1173,0.1216,0.0320,+0.483
[... 56 further rows, one per crown ...]

[... the tables per link and per ring of D1, D2 and D3, as for D0 in the opening message ...]

Their full fields are now under /data/designs (D1, D2, D3).

Simulations left: 11.
\end{Verbatim}

\begin{Verbatim}
best_design:
D13

summary:
D13 is the best feasible design from all 14 simulations. Design file: /work/tail_refine.json; key: d38c6420a4b5.

Verified results: Q = 0.108489, a 32.2% reduction from D0; inner mean diameter after recoil = 21.2521 mm; shortening = 1.2015 mm; volume = 228.5469 mm³; as-cut length = 39.1309 mm. Maximum crown strain is 0.112744 and maximum element PEEQ is 0.114017, with no elements above 0.1766. All geometry and feasibility checks passed.

The design uses 0.965666 mm shared width, 0.940 mm legs, D0's crown offset and ring pitch, and 1.500 mm link waist flats. Crown-specific profiles combine outer-apex reinforcement and shoulder relief on unlinked crowns, broader inner-edge reinforcement on linked crowns, and small asymmetric corrections informed by the element fields. Terminal extensions were removed.

The largest improvement came from spreading crown bending, not simply thinning struts. Longer legs reduced strain but consumed recoil-diameter margin; excessive shoulder relief created new strain peaks. Support-node changes also shifted the hottest crowns, limiting the predictability of fine local adjustments. D14 remained feasible but had slightly worse Q, so D13 is retained. Its diameter margin is only about 0.003 mm; these simulation results are not a fracture, fatigue, or clinical validation.

wanted_but_not_expressible:
Independent leg-width/curvature control, noncircular crown centerlines, curved link centerlines or different attachment topology, and through-thickness edge fillets were not available. For validation, I would also want mesh refinement and controlled support/seam sensitivity studies with consistent support mapping; those simulation controls were not exposed.
\end{Verbatim}

\paragraph{Other runs.}
The plate's sessions had the same system prompt, tools and records, and an opening message with the same seven
sections for the plate: the hole's two semi-axes and up to eight bumps, the simulation set-up, the original
design's peaks and its stress along the hole's edge every $3.75^\circ$ for both load cases, and the area limit.
Each design is submitted with predictions of the objective and of both load cases' peaks and their angles. One
rule is added: ``Do not solve the plate problem numerically yourself: no finite-element, boundary-element,
finite-difference or similar numerical solution of the plate, with your own code or with a library's. The
simulations of your budget are the only solutions of the plate in this session.'' A scan of every Python
program for code that could solve the plate flagged 8 of 505 programs on the judge's model and 5 of 241 on
Tosca's; read one by one, none solves it. On Tosca's coarse model, the opening message also describes the judge's
fine model and gives the start design's value there; the agent saw no other value of the judge during its
session.

The crown's sessions received their framework's own instructions as the system prompt and a task in four steps:
have a critic sub-agent review the fixed protocol and file its verdict; submit designs in rounds of up to three,
each with a reason and predictions of the peak strain, the opening and the shortening; read every design's
results before the next round, among them where along the crown the plastic strain sits; and finish with the
best design, a summary and what could not be expressed with the four values. Every crown session but the first
was also told that the protocol is fixed for the whole session and must not be edited or replaced. The task
ended:
``Nobody will answer questions during this session: decide yourself.'' The agent had a design-study tool, through
which it binds the review, submits rounds, reads the results and finishes, a tool that puts the critic's review
on record, and file tools confined to the session's folder; it had no shell and no Python sandbox. The set-up
sessions got the bar of \cref{app:crown} in plain words and, besides these tools, three more for looking up the
solver's documentation, retrieving example inputs and checking an input before it runs.

The pilot sessions led up to the clean run. The single call and the round on the shared settings had no tools:
their system prompt asked for one JSON object, and their opening message gave the same kind of facts, with
questions to answer and values to predict. Of the single call's three variants, one left out the tables per
crown, link and ring and kept the global results; one kept every number and table but replaced every physical
description by neutral names, such as parameters p1 to p7, ``units'' and ``fixed points''; and one repeated the
original call. The crown-by-crown session added the tools, with one design per simulation, and the
edge-profile session added the profiles and rounds of up to three. The shuffled-results control continued the
clean run's own conversation from the results of a round, with the same tools, sandbox and checks, and ended each
continuation at its first accepted round.

\paragraph{Records and reproduction.}
Every run's record holds its opening message as sent, every message, tool call and result, the tokens and costs
of its LLM calls, the images the agent viewed and the files it wrote, and, for every simulated design, its
design file, geometry, mesh and solver input, the exported fields, the digest and the score. On the stent and
the plate, the driver records each step before taking the next, so a run that stops, for a pause, for solver
licences or for a lost connection, resumes from its record without repeating an LLM call or a simulation. Before the clean run, its driver ran two rounds
with a scripted stand-in for the LLM against a stand-in for the simulation, with a pause and a resume, and its
opening message was checked to hold nothing from the pilot sessions. LLM calls cannot be repeated exactly,
since the LLMs sample their replies and their providers may change them, so a rerun under the same protocol is
a new run; this is why the protocols fix in advance what counts across repeated runs (\cref{sec:pipeline}). The
records reproduce every score of the agent's designs, which the judge computes from the stored fields, and they
allow a run to be continued from a recorded point. The shuffled-results control did so: it restarted the clean
run's conversation from the results of a round, after rebuilding the agent's working folder by running its
recorded Python programs again, each of which had to print what it had printed before.

\section{Stent finite-element model}\label{app:stent-model}

\paragraph{Geometry and load case.}
The original design is cut from a tube of 15.99\,mm outer diameter and 0.593\,mm wall thickness and is
38.74\,mm long; its struts are 0.963\,mm wide and narrow to 0.785\,mm at the crown tips. A rigid cylindrical
surface acts as the balloon: it is driven radially to a diameter of 21.6\,mm and back, expanding the stent
from its as-cut state and then letting it recoil freely, with hard contact and friction 0.05 between balloon
and stent and frictionless self-contact. The analysis is static with large deformations in Abaqus/Standard
2022.

\paragraph{Material.}
The magnesium alloy Mg-4Y-3Gd from the public data of \citet{steglich2025smartrim} is modelled with linear
elasticity ($E=43$\,GPa, $\nu=0.3$) and von Mises plasticity with isotropic hardening from one monotonic
tensile test, from 102\,MPa at first yield to 247.5\,MPa at an equivalent plastic strain of 0.1766, where the
data end; beyond that the stress is held constant.

\paragraph{Supports and mesh.}
Three nodes on the middle ring, $120^\circ$ apart and about 0.09\,mm below its centre plane, are fixed axially
and tangentially; they stop rigid-body motion but load the nearby crowns artificially. Moving them to another ring
lowered the original's most strained crown by 10\% but $Q$ by only 1.4\%. Abaqus/CAE meshes each
design with eight-node hexahedra (C3D8), swept through the wall in three layers with an in-plane seed of
0.19\,mm, about 37\,000 elements.

\paragraph{Crown strains.}
The element strain is the equivalent plastic strain (PEEQ) at the end of recoil, the largest of the element's
eight integration points. A per-crown trace assigns every element to its nearest crown along the design's own
outline, so a crown includes its arc and the adjoining half-legs, and a crown's strain is the mean of its five
highest element values.

\paragraph{Rebuild noise.}\label{app:noise}
Every new design gets its own CAD solid, a new free mesh and supports placed anew by the rule above, so its score
scatters in a way that re-solving the same mesh does not show. We measured the scatter by rebuilding four
designs with their struts 0.001\,mm thinner, a change expected to lower $Q$ by about 0.0002. All four rebuilds
scored higher than their originals, by 0.0008 to 0.0031, as expected when the originals were the best designs
of their campaigns, because the best of several noisy evaluations is optimistic. The scatter of $Q$, the root
mean square of the four changes divided by $\sqrt{2}$, is 0.0015; the difference between two rebuilds scatters
by $\sqrt{2}$ times as much, and we take twice that, 0.004, as the noise of a rebuild. The most strained crown
scatters by 0.006, about four times as much as $Q$, so its noise of a rebuild would be about 9\% of the
original's value, against 2.5\% for $Q$. The lumen scatters by
0.0005\,mm, well inside its tolerance. The shortening scatters by 0.06\,mm, twice its tolerance, because it
rests on the extreme nodes at the stent's ends, and one rebuilt design went from meeting the shortening limit
to breaking it; its noise, computed like that of $Q$, is 0.17\,mm, the margin of the head-to-head reading
(\cref{sec:objective}).

\paragraph{Computing.}
Our simulations ran in Abaqus~2022 on Strand, the HPC cluster of Helmholtz-Zentrum Hereon. Each design of
the agent was meshed and checked in about three minutes and solved on one core of a node with two 24-core
Intel Xeon Platinum 8160 processors at 2.1\,GHz. The clean run's fifteen solves took 1.86 to 2.29\,h each,
2.14\,h in the median and 31.4 core-hours in all; they mostly ran one after another, because the solver
licences, which we share with colleagues, rarely left room for two, and with the agent's own time and these
waits the run took 35\,h from its first solve to its last. Tosca solved its designs on three or four cores, in
1.1 to 1.7\,h each. The full-cycle model, as recorded in its solver's job summaries, took 10.1\,h for crimping
and the crimper's release and 7.0\,h for the expansion and the balloon's release of the original design, both
on two cores, 17.1\,h in all and 33 core-hours; for the agent's pilot design it took 25.0\,h on two cores and
14.2\,h on one, 39.2\,h in all and 63 core-hours. A run of the full chain thus costs 16 to 29 times the
core-hours of a solve of the reduced model.

\section{The plate with a hole}\label{app:plate}

\paragraph{Problem.}
Tosca's bundled shape example, copied unchanged from the Abaqus 2023 installation, is a quarter of a
$100\times100$\,mm plate, 1\,mm thick, with a central hole of radius 20\,mm, in plane stress (CPS4,
$E=210$\,GPa, $\nu=0.3$) with symmetry supports on the two cut edges. Two load cases are solved separately: one
pulls the right edge with 1800\,N in all, the other the top edge with 900\,N. Tosca's task is to lower the
largest von Mises stress on the hole's edge with its controller-based method, moving the edge nodes of a mesh
with 16 elements around the quarter hole. Our test reproduced the vendor's run: 162.30 to 109.22\,MPa in five
cycles. For every method alike we added a fine mapped mesh of the same type, with 96 elements around the
quarter hole, so that every design has the same mesh topology and no remeshing noise; the loads as uniform
tractions with the shipped totals; and a limit on the material area, at most the original's.

\paragraph{Judge and solver.}
The objective is Tosca's own measure, the largest von Mises stress at the hole's edge nodes over both load
cases, read on the fine mesh: 171.6\,MPa for the circle. Our reference solver implements the same element and
nodal stresses as Abaqus and reproduces Tosca's values on the shipped deck to all printed digits, and on all
15 meshes of Tosca's run to better than one part in ten million. Every agent
run's best design was also solved on a mesh twice as fine, where it reads at most 1.6\% higher and the order of
the runs is unchanged. Tosca's best design reads 1.8\% higher there (111.4\,MPa against 109.4), enough to fall
behind one more agent run.

\paragraph{Geometry parametrisation and agents.}
The geometry parametrisation, built on the stent's geometry code, sets the hole's two semi-axes (10 to 35\,mm) and up
to eight bumps on its edge (centre 0 to $90^\circ$, half width 10 to $90^\circ$, amount $-6$ to 6\,mm),
mirrored at the symmetry lines; the outline is extruded instead of wrapped. Rules keep the edge 5\,mm from the
outer edges and the hole's centre and the mesh's element angles within 15 to $165^\circ$. The agents had the
stent's tools and budget. One rule the stent did not need: an agent must not solve the plate numerically
itself, since a linear 2D solve is easy to write; an audit of every python cell found none that did.

\begin{figure}[t]
  \centering
  \includegraphics[width=\linewidth]{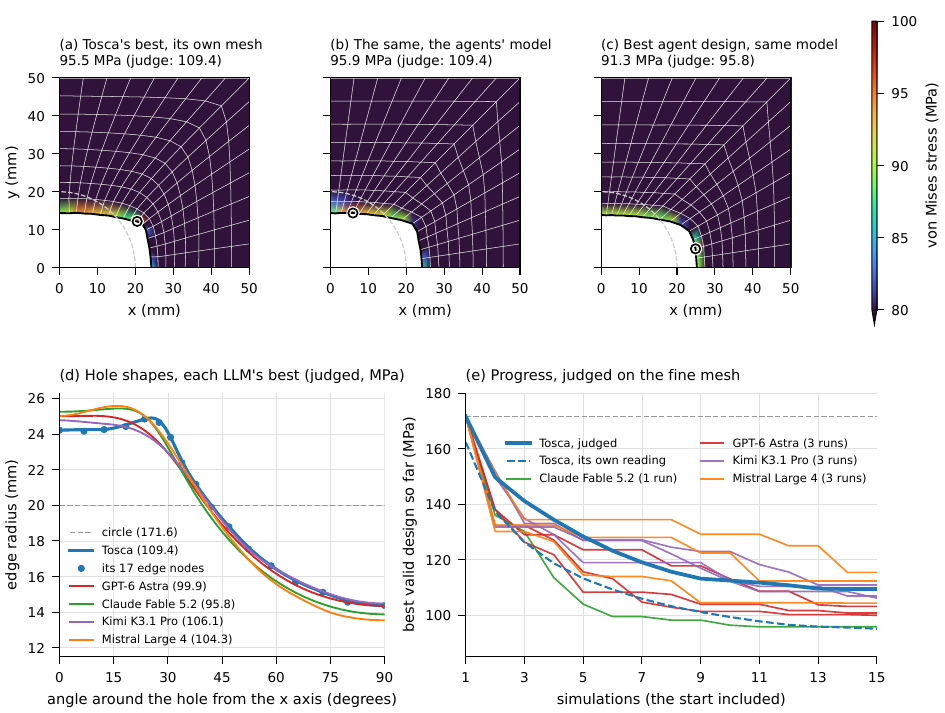}
  \caption{The plate on Tosca's own coarse model, with 16 elements along the quarter hole. (a)~Tosca's best
  design within 15 simulations, as its own mesh with its own loads. (b)~The same design on the coarse model the
  agents were given: the same 17 edge nodes, with the shipped mesh's straight lines inside where Tosca smooths
  them, which changes the stress at the hole by at most 1.1\%. (c)~The best agent design on that model, by
  Claude Fable~5.2. Colour: von Mises stress, the larger of the two load cases; circle: the edge node with the
  highest stress, where each value is read; dashed: the original hole. (d)~The hole shapes: Tosca's 17 edge
  nodes and the curve the judge reads through them, and each LLM's best design from its runs on the coarse
  model, with the judge's value in brackets. (e)~The best valid design so far against simulations for every run
  on the coarse model, judged on the fine mesh, and Tosca's reading on its own mesh.}
  \label{fig:plate}
\end{figure}

\paragraph{Results.}
\Cref{tab:plate} lists every method. Given Tosca's own coarse model, with the same mesh, loads and budget as Tosca, seven of ten agent runs
still beat Tosca on the judge, every Astra run among them (\cref{fig:plate}); their results moved by 1.6 to 3.4\,MPa against their
fine-model runs, less than the spread between one LLM's runs (3 to 20\,MPa). Read on finer and finer meshes,
Tosca's best design rises from 95.5\,MPa on its own deck to 109.4 on the judge's mesh, so
its own model flatters it by 13\%, because it builds a sharp step near 25--$30^\circ$ (\cref{fig:plate}d) that
its 16 elements along the quarter hole cannot resolve; the coarse model flatters the agents' designs by 4 to
10\%. The agents knew how they would be judged, and one corrected for the finer mesh explicitly, which
Tosca cannot be told; their bumps are smooth by construction, so part of that robustness comes from the
parametrisation. The four best runs placed all eight bumps freely, tuning 24 to 26 values.

\begin{table}[t]
  \centering
  \caption{The plate with a hole: the best valid design within 15 solves, the start's included, on the
  judge's fine mesh (the circle: 171.6\,MPa). The agents changed the hole's two semi-axes and up to
  eight bumps on its edge, up to 26 values.}
  \label{tab:plate}
  \small
  \begin{tabular}{@{}lllr@{}}
    \toprule
    Method & Design space & Best per run (MPa) & Below the circle \\
    \midrule
    Claude Fable 5.2, three runs & Up to 26 values & 93.0, 92.7, 91.8 & 46--47\% \\
    GPT-6 Astra, three runs & Up to 26 values & 101.5, 102.6, 94.7 & 40--45\% \\
    Kimi K3.1 Pro, three runs & Up to 26 values & 103.7, 110.2, 115.0 & 33--40\% \\
    Mistral Large 4, three runs & Up to 26 values & 125.6, 107.4, 105.4 & 27--39\% \\
    Tosca, as shipped & Its mesh: 17 edge nodes & 109.4 & 36\% \\
    Bayesian optimisation & Two semi-axes & 125.9 & 27\% \\
    \bottomrule
  \end{tabular}
\end{table}

\section{Reduced crown problem: four values for every method}\label{app:crown}

On the stent, only the agent and Tosca can work on all of the design's values within the budget. A reduced
problem that
solves in a minute is the ladder's bottom rung (\cref{tab:ladder}), where the agent and the classical
optimisers change the same four variables. It answers three questions: whether reading the results helps
then, whether the result depends on the LLM, and what happens when the LLM must also write the
simulation set-up, which in the loop is written and checked by hand (\cref{tab:interface}). The sessions ran
in a lighter implementation of the protocol of \cref{sec:loop}, a design-study tool that we built on the
open-source agent framework openPASO~\citep{hermann2026oasis}.

\begin{figure}[t]
  \centering
  \includegraphics[width=\linewidth]{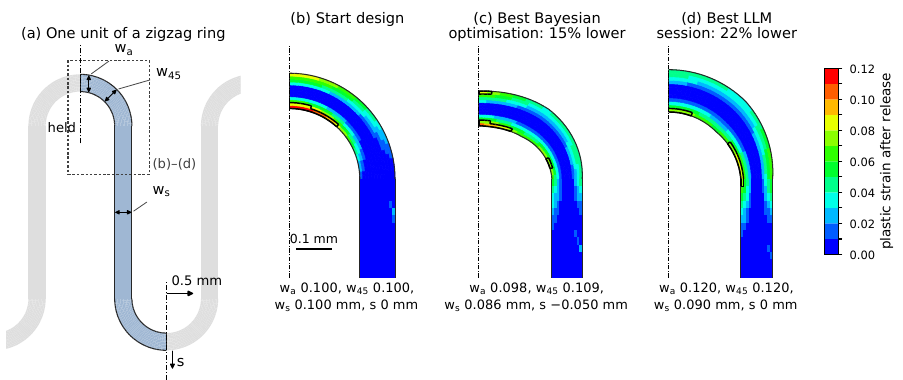}
  \caption{The reduced crown problem. (a)~The model: one unit of a zigzag ring, unrolled into the plane, from
  the middle of one crown down its strut to the middle of the opposite crown, with its neighbours mirrored in
  grey. The four values set the width at the apex, $w_a$, $45^\circ$ from it, $w_{45}$, and along the strut,
  $w_s$, and an outward shift $s$ of the apex; both crowns take the same values. Dash-dotted: the symmetry
  planes, one held, the other moved by 0.5\,mm and released. (b)--(d)~The plastic strain after release in the
  upper crown, the box in~(a); the lower crown mirrors it. (b)~The start design. (c)~The best design of Bayesian
  optimisation over its five seeds. (d)~The best LLM design, by Mistral Large~4; the best designs of
  GPT-6 Astra, Kimi K3.1 Pro and DeepSeek V4.1 Flash are alike, with a wider apex and shoulder and a thinner
  strut. Outlined: the most strained 2\% of the metal, whose mean strain is the objective; below each panel,
  its four values. Each design solved again from its own input, which reproduces its recorded result exactly.}
  \label{fig:crownmodel}
\end{figure}

\paragraph{Problem.}
One crown of a zigzag ring, unrolled into the plane: half a crown, its strut and half of the opposite crown,
with both cut faces on symmetry planes, opened by 0.5\,mm per half period and released (\cref{fig:crownmodel}a).
FEBio~4.12~\citep{maas2012febio} solves it with von Mises plasticity ($E=200$\,GPa, $\nu=0.3$, yield stress
300\,MPa, linear hardening 2\,GPa) on a structured mesh of 32, 48 and 12 elements along the arc, along the
strut and across the width, in one to two minutes on one core. Four variables set the width profile: the
width at the apex, $w_a$, $45^\circ$ from it, $w_{45}$, and along the strut, $w_s$, each between 0.06 and
0.2\,mm and 0.1\,mm at the start, and an outward shift $s$ of the apex of up to $\pm0.05$\,mm. The objective
is the mean plastic strain of the
most strained 2\% of the metal after release, 0.1003 for the start design. In checks before any session it
ranked five test designs as a mesh 1.5 times finer does, whereas a mean over the four most strained elements
moved by 30--50\% on a mesh twice as fine. A design is feasible if its opening after recoil is at least the
start design's and its shortening at most the start design's, each with a tolerance of 0.001\,mm fixed in
advance, and if its metal volume, exact in closed form, is at most the start design's. Every method gets
twelve simulations beyond the start design. The four parts a problem supplies (\cref{sec:pipeline}) are, for this problem: four
bounded numbers written into a script that generates the solver input; the FEBio model; a digest that tables
the plastic strain along the crown, edge by edge, next to the scores; and the objective and limits above. The
digest shows the agent where along the crown the strain sits, whereas the classical optimisers see only the
scores.

\paragraph{Agent and protocol.}
The agent submits rounds of up to three designs, each with a reason and predictions of the peak strain, the
opening and the shortening, and reads each design's digest and solver output as files. It has no Python
sandbox, no image viewer and no shell, and no simulation outside its budget; no CAD is built. Before the first
simulation, a critic sub-agent of the same LLM reviews the fixed protocol. The judge follows the rules described in Section~\ref{sec:pipeline}: a failed simulation counts and is never feasible, every limit is the start design's own
value, and every limit is read both exactly and with its tolerance. The protocol was fixed before the first
call; five dated amendments, each made before the sessions it governs, added sessions, raised caps and, after
the first session, told the agent not to edit the protocol.

\paragraph{Sessions and baselines.}
Twenty crown sessions used six LLMs: GPT-6 Astra (one session), Qwen~3.5~397B (three), Mistral Large~4
(three), Qwen~3.8 (one), and Kimi~K3.1~Pro and DeepSeek~V4.1~Flash (three each, and three more each with a
larger cap). In the round that added Mistral, Qwen~3.8, Kimi and DeepSeek, the caps were sized on input tokens;
Kimi and DeepSeek, which spend most of their money on reasoning, used them up early, mostly in the opening
review, so both were run again with US\$2 per session. Two Mistral sessions and the Qwen~3.8 session were also
stopped by their caps, after six to nine simulations, and were not run again. One Qwen~3.5 session followed its
critic's false objection to the exact volume formula, edited the protocol and left the study; it is listed but
not compared. The classical baselines start from the same design with the same twelve simulations and judge:
COBYLA with step sizes 0.1, 0.2 and 0.4 in scaled variables, random search with five seeds, and Bayesian
optimisation with five seeds. The Bayesian optimisation is the one we built for the stent and also ran on the
plate: BoTorch~\citep{balandat2020botorch}, one Gaussian process per output with a Mat\'ern-5/2 kernel, log
constrained expected improvement~\citep{ament2023logei}, and four Sobol points before eight surrogate-guided
steps. All three skip, without solving them, designs that break a rule on the widths, which the agent's free
check also shows. Bayesian optimisation and random search also skip designs with more metal than the start
design, while COBYLA treats the volume as one of its constraints and solves such designs. \Cref{tab:crown}
lists every session and baseline.

\begin{figure}[tbp]
  \centering
  \includegraphics[width=\linewidth]{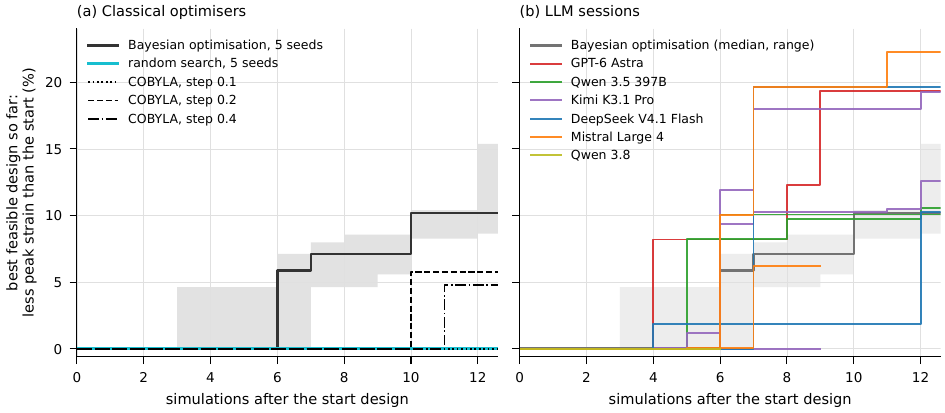}
  \caption{The reduced crown problem: the best feasible design so far against simulations, main reading, study
  mesh. (a)~The classical optimisers; random search and COBYLA with step size 0.1 found no feasible
  improvement. (b)~Every LLM session under the final caps for its LLM, except the one that left the
  study, over the median and range of Bayesian optimisation.}
  \label{fig:crown}
\end{figure}

\paragraph{Results.}
With the same four variables and twelve simulations, random search found no feasible design better than the
start in any of its five seeds, and COBYLA lowered the peak strain by 0 to 5.8\%, depending on its step size
(\cref{tab:crown,fig:crown}). Bayesian optimisation was the strongest baseline: 8.6 to 15.4\% over its five
seeds, with a median of 10.2\%. The nine LLM sessions that used all twelve simulations reached 10.1
to 22.3\%, so none fell clearly behind the median seed, and the best session of each of four LLMs, Mistral
Large~4 (22.3\%), DeepSeek~V4.1~Flash (19.7\%), GPT-6 Astra (19.3\%) and Kimi~K3.1~Pro (19.3\%), also beat
the best seed. On the mesh 1.5 times finer, the sessions' improvements changed by at most 1.5 percentage points
and those of Bayesian optimisation by at most 1.8, while one COBYLA lead did not hold (4.8\% fell to 0.4\%).
The ten sessions that a cost cap or empty replies ended early reached 0 to 6.2\%. The criterion fixed before
the baselines ran, a larger reduction than the median Bayesian-optimisation seed on both meshes in at least two
of an LLM's three sessions, is met by Kimi; DeepSeek, Mistral and Qwen~3.5 meet it once each, and GPT-6 Astra
and Qwen~3.8 ran once. Under the strict reading, with every limit met exactly, the difference grows: Kimi's best
design lowered the peak strain by 19.3\% and GPT-6 Astra's by 14.8\%, against at most 5.1\% for any classical
baseline (6.0\% on the finer mesh).

\paragraph{Paths of the best sessions.}
The sessions converged on the same trade-off: a stronger taper, wider at the apex and shoulder and narrower
along the strut, lowers the strain but costs opening after recoil, so most of the best designs sit on the
opening limit and use its tolerance. Where the start design's most strained metal is a patch on the inner
edge within $40^\circ$ of the apex, two elements deep, these designs spread it thinner and further along the
inner edge of the bend (\cref{fig:crownmodel}). GPT-6 Astra then solved the same widths with three shifts of the apex,
measured that an inward shift wins back opening for a little strain, and spent that opening on a stronger
taper; Kimi's best design makes the same move and needs no tolerance. Bayesian optimisation found this region
in one of its five seeds (15.4\%, with the apex at its inward bound); its other seeds thinned the shoulder
instead (8.6 to 10.4\%). The LLMs' predictions of the peak strain sharpened as they read the results: in seven
of the nine complete sessions, their error fell from 6 to 100\% of the start design's value in the first round
to 0.1 to 7\% in the last, about as accurate as the Gaussian processes' predictions in Bayesian optimisation's
later steps (2 to 9\%).

\begin{table}[tbp]
  \centering
  \caption{Every crown session and classical baseline, each from the same start design with twelve
  simulations: the peak strain of the best feasible design, in percent below the start design's, under the
  main reading (the tolerances fixed in advance) and the strict reading (every limit exact), on the study mesh
  and on a mesh 1.5 times finer, where the design is solved again and judged against the start design on that
  mesh; 0.0 means that no feasible design beat the start. Cap and cost: US\$ per session at list prices, helper
  agents included; ``cap'': ended by its cost cap; ``no answer'': ended by empty replies. For Kimi and
  DeepSeek, the second group of three ran with the larger cap.}
  \label{tab:crown}
  \footnotesize
  \begin{tabular}{@{}llrlrrrrr@{}}
\toprule
& & & & \multicolumn{2}{c}{Main reading} & \multicolumn{2}{c}{Strict reading} & \\
\cmidrule(lr){5-6}\cmidrule(lr){7-8}
Method & Run & Cap & Simulations & Study & Finer & Study & Finer & Cost \\
\midrule
COBYLA & step 0.1 & -- & 12 & 0.0 & 0.0 & 0.0 & 0.0 & -- \\
 & step 0.2 & -- & 12 & 5.8 & 5.6 & 0.0 & 0.0 & -- \\
 & step 0.4 & -- & 12 & 4.8 & 0.4 & 0.0 & 0.0 & -- \\
\addlinespace
Random search & seed 0 & -- & 12 (1 failed) & 0.0 & 0.0 & 0.0 & 0.0 & -- \\
 & seed 1 & -- & 12 (1 failed) & 0.0 & 0.0 & 0.0 & 0.0 & -- \\
 & seed 2 & -- & 12 (1 failed) & 0.0 & 0.0 & 0.0 & 0.0 & -- \\
 & seed 3 & -- & 12 (2 failed) & 0.0 & 0.0 & 0.0 & 0.0 & -- \\
 & seed 4 & -- & 12 & 0.0 & 0.0 & 0.0 & 0.0 & -- \\
\addlinespace
Bayesian optimisation & seed 0 & -- & 12 (1 failed) & 10.2 & 10.4 & 0.0 & 0.0 & -- \\
 & seed 1 & -- & 12 (1 failed) & 10.0 & 9.3 & 0.0 & 0.0 & -- \\
 & seed 2 & -- & 12 & 8.6 & 6.8 & 4.6 & 6.0 & -- \\
 & seed 3 & -- & 12 & 10.4 & 10.9 & 0.0 & 0.0 & -- \\
 & seed 4 & -- & 12 & 15.4 & 15.3 & 5.1 & 5.3 & -- \\
\addlinespace
GPT-6 Astra & 1 & 30 & 12, cap & 19.3 & 20.8 & 14.8 & 16.3 & 38.39 \\
\addlinespace
Qwen 3.5 397B & 1 & 3 & \multicolumn{5}{l}{left the official study} & 0.38 \\
 & 2 & 3 & 12 & 10.6 & 10.9 & 0.0 & 0.0 & 0.40 \\
 & 3 & 3 & 12 & 10.1 & 10.6 & 0.0 & 0.0 & 0.39 \\
\addlinespace
Mistral Large 4 & 1 & 1 & 9, cap & 6.2 & 5.2 & 0.0 & 0.0 & 0.94 \\
 & 2 & 1 & 12 (1 failed) & 22.3 & 22.0 & 0.0 & 0.0 & 0.84 \\
 & 3 & 1 & 6, cap & 0.0 & 0.0 & 0.0 & 0.0 & 0.92 \\
\addlinespace
Qwen 3.8 & 1 & 2.5 & 6, cap & 0.0 & 0.0 & 0.0 & 0.0 & 2.25 \\
\addlinespace
Kimi K3.1 Pro & 1 & 0.75 & 12, cap & 10.3 & 9.9 & 5.7 & 4.8 & 0.67 \\
 & 2 & 0.75 & 0, cap & -- & -- & -- & -- & 0.72 \\
 & 3 & 0.75 & 3, cap & 0.0 & 0.0 & 0.0 & 0.0 & 0.67 \\
 & 1 & 2 & 9, cap & 0.0 & 0.0 & 0.0 & 0.0 & 2.31 \\
 & 2 & 2 & 12, cap & 12.6 & 12.4 & 8.7 & 8.3 & 1.93 \\
 & 3 & 2 & 12 & 19.3 & 18.4 & 19.3 & 18.4 & 1.82 \\
\addlinespace
DeepSeek V4.1 Flash & 1 & 0.25 & 0, cap & -- & -- & -- & -- & 0.36 \\
 & 2 & 0.25 & 0, cap & -- & -- & -- & -- & 0.28 \\
 & 3 & 0.25 & 0, cap & -- & -- & -- & -- & 0.24 \\
 & 1 & 2 & 12 (1 failed) & 10.3 & 10.0 & 0.0 & 0.0 & 1.33 \\
 & 2 & 2 & 6, no answer & 0.0 & 0.0 & 0.0 & 0.0 & 1.34 \\
 & 3 & 2 & 12 (3 failed), cap & 19.7 & 20.5 & 2.1 & 1.5 & 2.11 \\
\bottomrule
\end{tabular}

\end{table}

\paragraph{Study set-up from a plain description.}
In twenty further sessions the LLM received a plain-language task and had to write the whole study itself:
the solver input, the script that reads the results, the objective and the limits. The task was a bar of five
linearly varying widths, pulled with plasticity, to strain as little as possible with no more metal and no less
pulling force; its optimum, a uniform bar, is known. GPT-6 Astra, Kimi (three of three sessions), Mistral
(three of three), DeepSeek (two of three) and Qwen~3.8 (one of one) set up a study that measured the actual
task, and all but one of those ten sessions found the optimum. Qwen~3.5, which improved the crown when it was
given the set-up, managed this in one of nine sessions. Five of the others never got their start design to
run, three of them before we fixed how the solver's error messages reached the LLM, and three accepted a
set-up that ran but measured something else, an elastic material instead of the plastic one, a pulling force
of zero or no strain at all, and reported success. Writing the set-up is where the weaker LLM failed, which
is why, in the loop, the four parts of a problem are written and checked before any call and the agent only
changes the design.

\paragraph{Limits.}
The problem is small: four variables and a known trade-off between strain and opening. GPT-6 Astra ran once.
The caps differed between rounds. The implementation differs from the stent's: no sandbox, no CAD, and a critic
before the first simulation. FEBio on this fixed, structured mesh is deterministic, so the finer mesh, not a
rebuild, is the yardstick for every gap.

\bibliographystyle{plainnat}
\bibliography{references,references-capability}

\end{document}